\documentclass[twoside,11pt]{article}
\usepackage{jair, theapa, rawfonts}
\usepackage{latexsym, lipsum}
\usepackage[ruled,linesnumbered]{algorithm2e}
\usepackage{amsmath,amssymb,graphicx,wasysym,bbm}
\usepackage{multirow,booktabs,xcolor,colortbl}
\usepackage{tabularx}
\usepackage[caption=false]{subfig}
\graphicspath{{figures/}}

\SetLipsumParListSurrounders{\colorlet{oldcolor}{.}\color{gray}}{\color{oldcolor}}

\jairheading{74}{2022}{1059-1089}{12/2021}{07/2022}
\ShortHeadings{
Supervised Visual Attention for Simultaneous MMT}
{Haralampieva, Caglayan, \& Specia}
\firstpageno{1059}

\providecommand\longrightarrowRHD{\relbar\joinrel\relbar\joinrel\mathrel\RHD}
\providecommand\longrightarrowrhd{\relbar\joinrel\relbar\joinrel\mathrel\rhd}

\makeatletter
\providecommand*\xrightarrowRHD[2][]{\ext@arrow 0055{\arrowfill@\relbar\relbar\longrightarrowRHD}{#1}{#2}}
\providecommand*\xrightarrowrhd[2][]{\ext@arrow 0055{\arrowfill@\relbar\relbar\longrightarrowrhd}{#1}{#2}}
\newcommand*{\@rowstyle}{}
\newcommand*{\rowstyle}[1]{
  \gdef\@rowstyle{#1}%
  \@rowstyle\ignorespaces%
}
\newcolumntype{=}{
  >{\gdef\@rowstyle{}}%
}

\newcolumntype{+}{
  >{\@rowstyle}%
}
\makeatother

\def\ie{i.e.\ }
\def\eg{e.g.\ }
\newcommand\ra{$\rightarrow$}
\newcommand{\MC}[3]{\multicolumn{#1}{#2}{#3}}
\newcommand{\MR}[3]{\multirow{#1}{#2}{#3}}

\newcommand{\B}{\textbf}
\newcommand{\I}{\textit}
\newcommand{\T}{\texttt}

\newcommand{\nmt}{\textsc{Baseline Nmt}}
\newcommand{\mmt}{\textsc{\,+ Images}}
\newcommand{\mmtsup}{\textsc{\,\,\,+ Sup (Scratch)}}
\newcommand{\mmtsupf}{\textsc{\,\,\,+ Sup (Fine-Tuned)}}

\newcommand{\Tf}[1]{\textbf{#1}}
\newcommand{\Tt}[1]{#1}

\newcommand{\G}[1]{\textcolor{gray}{#1}}
\newcommand{\C}[1]{\textcolor{blue}{#1}}
\newcommand{\W}[1]{\textcolor{red}{#1}}

\begin{document}

\title{Supervised Visual Attention for Simultaneous\\Multimodal Machine Translation}

\author{\name Veneta Haralampieva
        \email veneta.l.haralampieva@gmail.com \\
        \name Ozan Caglayan {\normalfont (Corresponding author)}
        \email o.caglayan@ic.ac.uk \\
        \name Lucia Specia
        \email l.specia@ic.ac.uk \\
        \addr Department of Computing, Imperial College London, UK}
        
\maketitle

\begin{abstract}
There has been a surge in research in multimodal machine translation (MMT), where additional modalities such as images are used to improve translation quality of textual systems. A particular use for such multimodal systems is the task of simultaneous machine translation, where visual context has been shown to complement the partial information provided by  the source sentence, especially in the early phases of translation. In this paper, we propose the first Transformer-based simultaneous MMT architecture, which has not been previously explored in simultaneous translation. Additionally, we extend this model with an auxiliary supervision signal that guides the visual attention mechanism using labelled phrase-region alignments. We perform comprehensive experiments on three language directions and conduct thorough quantitative and qualitative analyses using both automatic metrics and manual inspection. Our results show that (i) supervised visual attention consistently improves the translation quality of the simultaneous MMT models, and (ii) fine-tuning the MMT with supervision loss enabled leads to better performance than training the MMT from scratch. Compared to the state-of-the-art, our proposed model achieves improvements of up to 2.3 BLEU and 3.5 METEOR points.
\end{abstract}

\section{Introduction}
\label{sec:intro}
Simultaneous machine translation (MT) aims at providing a computational framework that 
reproduces how human interpreters perform \I{simultaneous interpretation}. In this task, the duty of the interpreter is to translate speech in near real-time, by constantly maintaining a balance between the time needed to accumulate sufficient context and the translation \I{latency} the listeners experience in return.
This \I{streaming} property is what differentiates simultaneous MT from the conventional MT approaches, which process complete source \I{sentences}. Traditional work in simultaneous translation have dealt with this \I{streaming} property by relying on syntactic or heuristic constraints~\shortcite{bub1997verbmobil,ryu-etal-2006-simultaneous,bangalore-etal-2012-real} to determine the amount of \I{waiting} prior to \I{committing} a partial translation. Similar approaches have also been explored using state-of-the-art neural MT (NMT) architectures~\shortcite{sutskever2014sequence,Bahdanau2015NeuralMT,vaswani2017attention}, such as rule-based deterministic policies implemented at (greedy) decoding time~\shortcite{cho2016can} or the \textsc{wait-k} policy which sequentially switches between reading a new word and committing a translation~\shortcite{ma2019stacl}. Adaptive policies, which attempt to learn when to \I{read} or \I{commit}
depending on the context, have also been explored mostly through reinforcement learning-based techniques~\shortcite{gu2016learning,alinejad2018prediction,ive-etal-2021-exploiting}.

In this work, we focus on the \I{translation quality} aspect of the simultaneous translation framework and explore whether input contexts other than the linguistic signal
can improve the performance of simultaneous MT models.
Although such additional information may naturally occur in human simultaneous interpretation through the \I{a priori} knowledge of factors such as the topic, speaker or even the venue of the speech, in a computational model any additional context should be explicitly and carefully integrated to explore different inductive biases during model learning. Therefore, to mimic the availability of multiple input modalities for simultaneous MT, we follow the multimodal machine translation (MMT) framework where the objective is to translate image descriptions into different languages, by integrating the images themselves as additional context~\shortcite{specia-etal-2016-shared,sulubacak2020multimodal}. Intuitively, the expectation is that as long as there is a correlation between the language and the visual semantics, this way of grounding language can help anticipate future context for better translation quality, and even reduce the \I{latency} for adaptive policies.

Although relatively few, several works have explored a similar framework to analyse the benefits of visual grounding to simultaneous MMT: \shortciteA{caglayan2020simultaneous} and \shortciteA{imankulova-etal-2020-towards} approached the problem by integrating visual features into the encoder and/or the decoder of recurrent MMT architectures and coupling them with deterministic \textsc{wait-k} policy~\shortcite{ma2019stacl} and rule-based decoding algorithms~\shortcite{cho2016can}. \shortciteA{ive-etal-2021-exploiting} introduced reinforcement learning (RL)-based adaptive policies for recurrent MMT architectures.

In this paper, we first propose a Transformer-based~\shortcite{vaswani2017attention} simultaneous MMT model where regional features extracted from a state-of-the-art object detector~\shortcite{Anderson2017up-down}, are fused with the source language representations using an attention based cross-modal interaction (CMI) layer. To implement simultaneous multimodal translation, we adopt the \I{prefix training} approach~\shortcite{niehues2018,arivazhagan2020re} and evaluate its performance along with the deterministic \textsc{wait-k} policy~\shortcite{ma2019stacl}. Next, we propose a novel approach to simultaneous MMT to improve the grounding -- and therefore the anticipation -- ability of our model. The proposed method involves \I{supervising} the alignment between the source language representations and the image regions, through the use of labelled phrase-region correspondences.
We devise two multi-task learning settings to enable the visual supervision: (i) fine-tuning a pre-trained MMT for a fixed number of epochs, or (ii) training the MMT from scratch. We perform extensive experiments on the three different language pairs (English\ra\{Czech,French,German\}) of the Multi30k dataset~\shortcite{elliott-etal-2016-multi30k} and conduct
thorough quantitative and qualitative analyses to understand the potential impacts of attention supervision. Our results show that (i) \I{prefix training} achieves substantially better scores than the \textsc{wait-k} approach, (ii) supervised visual attention consistently improves the translation quality of the MMT models, and (ii) fine-tuning the MMT offers better performance than training the MMT from scratch. Finally, our supervised models achieve up to 2.3 BLEU~\shortcite{papineni2002bleu} and 3.5 METEOR~\shortcite{meteor} points improvements over the current state-of-the-art on Multi30k dataset.

The remaining of the paper is organised as follows: we provide a detailed account of literature in \S~\ref{sec:related} and describe our methodology and resources in \S~\ref{sec:method}. In \S~\ref{sec:results}, we present the quantitative and qualitative results across different language pairs and test sets of Multi30k. Finally, we conclude our work in \S~\ref{sec:conclusion} with a discussion on possible directions for future research.

\section{Related Work}
\label{sec:related}
\begin{figure}[t!]
\centering
\includegraphics[width=0.98\textwidth]{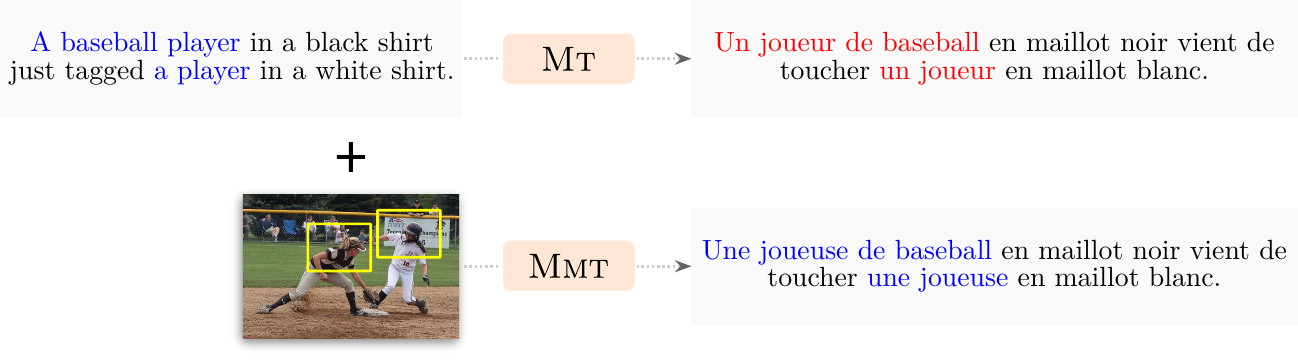}
\caption{Multimodal machine translation (MMT) can help disambiguate the source sentence when translating into gender-marked languages. The example is taken from the Multi30k dataset~\shortcite{elliott-etal-2016-multi30k}.}
\label{fig:mmt}
\end{figure}

The primary goal of Multimodal Machine Translation (MMT) is to improve the quality of machine translation (MT) by incorporating information from additional sources such as images or videos~\shortcite{sulubacak2020multimodal}. Of these two approaches, image-guided MMT (Figure~\ref{fig:mmt}) is substantially more researched than the video-guided MMT, simply due to the availability of more training resources. Since this paper heavily relies on image-guided MMT, we begin this section with a detailed literature overview on MMT first, and then continue with simultaneous MT.

\subsection{Multimodal Machine Translation}

\paragraph{Attentive models.} Inspired by the success of the textual attention~\shortcite{Bahdanau2015NeuralMT} in NMT models, a considerable amount of research has focused on coupling visual attention and textual attention altogether to perform image-guided MMT. These works generally encode the images using state-of-the-art CNN models pre-trained on large computer vision datasets such as ImageNet~\shortcite{deng2009imagenet}. This way, an image is represented as a set of convolutional feature maps of size $K\times K\times C$ \ie  each output channel $C_i$ encodes activations across a uniformly partitioned $K\times K$ grid. Using these features, \shortciteA{caglayan2016does,calixto2017doubly}
explore shared and separate attention mechanisms to compute a cross-modal latent representation in the multi-modal space. In addition to the language context computed by the textual attention, the translation decoder is now also conditioned on the multi-modal representation. As a follow-up, \shortciteA{libovicky2017attention} proposed various extensions,  of which the hierarchical attention variant gained popularity. This method weighs the relevance of textual and visual contexts using a third attention mechanism, instead of simpler fusion strategies such as addition or concatenation.
\shortciteA{libovicky2018input} later adapted these extensions to Transformer-based~\shortcite{vaswani2017attention} NMT models to enable image-guided Transformer MMTs. Although the majority of the approaches in attentive MMT rely on decoder-side visual attention, encoder-side grounding was also explored, but only for recurrent architectures: \shortciteA{delbrouck2017modulating} propose adding a visual attention layer in the encoder, where
its states act as the \I{query} and the visual features are taken as \I{keys} and \I{values}.

\paragraph{Simple grounding.} Finally, another line of work investigates the use of \I{pooled} visual representations ($\in \mathbb{R}^C$) for image-guided MMT, instead of the dense convolutional feature maps.
These approaches usually initialise the hidden state of the encoder and/or the decoder in the model, using a projection of the visual feature vector~ \shortcite{calixto2016dcu,calixto2017incorporating,caglayan2017lium}. Multi-task learning is also explored~\shortcite{elliott-kadar-2017-imagination,zhou-etal-2018-visual} using these simple vectorial representations, where the model is tasked with the reconstruction of visual features using the encoder's output.

\subsection{Supervision of Attention}
Previous work has explored whether supervising the encoder-decoder attention can improve the alignment and translation quality of text-only NMT systems. \shortciteA{liu-etal-2016-neural,mi-etal-2016-supervised} investigate supervising the attention of a recurrent NMT model by adding an alignment loss which is jointly optimised alongside the negative log-likelihood objective. \shortciteA{garg2019jointly} later extend this to Transformers-based text-only NMT models, showing that supervising a single attention head from the cross-attention layers of the decoder, can outperform existing alignment models without a significant degradation in translation quality. 

Recently, \shortciteA{nishihara-etal-2020-supervised,specia2020read-spot-translate} investigate a multimodal co-attention mechanism \shortcite{lu2016hierarchical} in the encoder, which first uses an affinity matrix to capture the relationship between the source tokens and image features and then computes the visual and textual attention weights. \shortciteA{specia2020read-spot-translate} explore the impact of supervising the visual attention weights using ground-truth alignments obtained from the Flickr30k Entities dataset~\shortcite{plummer-etal-2015-collecting} and show that their multimodal systems are better at disambiguating words, compared to their text-only baseline system. For \shortciteA{nishihara-etal-2020-supervised}, the improvements are marginal unless cross-language alignments are also supervised in addition to the visual attention.

\subsection{Simultaneous Machine Translation}
Early works in simultaneous neural MT (SiMT) explore using a pre-trained full-sentence NMT model at test time, by relying on specific decoding approaches designed for simultaneous interpretation. Of these, \shortciteA{cho2016can} propose a greedy decoding algorithm with two different waiting criteria based on simple heuristics which determine whether the model should \textit{READ} a source token or \textit{WRITE} a target one. Rather than relying on hand-crafted heuristics, several works investigate using Reinforcement Learning (RL) to learn an adaptive policy which maximises the translation quality and minimises the delay/latency.
\shortciteA{satija2016simultaneous} train an agent using Deep Q-Learning while \shortciteA{gu2016learning} rely on the policy gradient algorithm~\shortcite{williams1992simple}. \shortciteA{alinejad2018prediction} later extend the latter by adding a \textit{PREDICT} action, which enriches the available context by utilising predictions of future source tokens.   

A drawback of the approaches discussed so far is the discrepancy between training and test times of the underlying NMT model: the model being trained on full-sentence source contexts, are later exposed to partial contexts at test time. \shortciteA{dalvi2018incremental} propose mitigating this by fine-tuning the model using either chunked segments or prefix pairs. Next, \shortciteA{ma2019stacl} explore a fixed \textsc{wait-k} policy which can be used at both training and test times, with the model initially reading $k$ source tokens before proceeding to alternate between writing a single target token and reading a source one. Later, \shortciteA{arivazhagan2019monotonic} extend this to an adaptive policy using an advanced attention mechanism in a Recurrent NMT model and a weighted latency loss, while \shortciteA{ma2019monotonic} further develop this for the multi-head attention used in Transformers. More recently, \shortciteA{arivazhagan2020re} investigate whether training a model on prefix pairs and re-translating previously emitted target words at decoding time improves translation quality or not. Their results show that augmenting the training data with prefix pairs can outperform the \textsc{wait-k} trained systems, with re-translation further increasing the quality. 

An alternative method for obtaining an adaptive policy is to train a policy model using Supervised Learning and ground-truth action sequences which \shortciteA{zheng2019simpler} propose generating using a pre-trained NMT Transformer model, with the ideal action being a \textit{WRITE} when the ground truth target word is ranked within the top $k$ next word candidates, or \textit{READ} otherwise. \shortciteA{arthur2021learning} rely on a statistical word alignment system to obtain the ground-truth actions and use them to jointly train the translation and policy models. 

\paragraph{Simultaneous MMT.}
Previous work in simultaneous MMT mostly rely on rule-based strategies~\shortcite{cho2016can,ma2019stacl} on the Multi30k dataset for recurrent MMT models. Of these approaches, \shortciteA{imankulova-etal-2020-towards} explored the \textsc{wait-k} policy using a recurrent MMT equipped with a hierarchical multimodal attention. Specifically, for each $k$, they first conduct a textual pre-training (\ie without the visual features) until convergence, and then fine-tune the checkpoint with visual features enabled.
\shortciteA{caglayan2020simultaneous} conducted a study where they compare object classification and object detection features for two different multimodal architectures: decoder-level visual attention and encoder-level visual attention. As for the simultaneous translation part, they explore both \textsc{wait-k} and rule-based decoding~\shortcite{cho2016can} methods. Finally, \shortciteA{ive-etal-2021-exploiting} attempted to learn a multimodal policy through reinforcement learning, for deciding the \I{ READ/WRITE} actions during simultaneous translation.

Our work resembles to \shortciteA{caglayan2020simultaneous} as
we explore the deterministic \textsc{wait-k} policy along with the object detection features extracted from salient regions. We also add another policy to our inventory and more importantly, we investigate the impact of supervising the visual attention using human-labelled annotations.

\section{Method}
\label{sec:method}
This section presents the approaches
explored in this work. We first begin with a description of the text-only Transformers NMT~\shortcite{vaswani2017attention} and how we extended it 
to accommodate simultaneous translation. Next, we describe our baseline multimodal Transformer architecture that incorporates visual attention. Finally, we introduce our approach to supervise the visual attention in simultaneous MMT.

In what follows,
the source sentence and target sentence tokens are denoted with $\mathbf{x} = [x_1, x_2, \dots, x_N]$ and $\mathbf{y} = [y_1, y_2, \dots, y_M]$, respectively.

\subsection{Transformer-based NMT}
\begin{figure}[t!]
\centering
\includegraphics[width=0.65\textwidth]{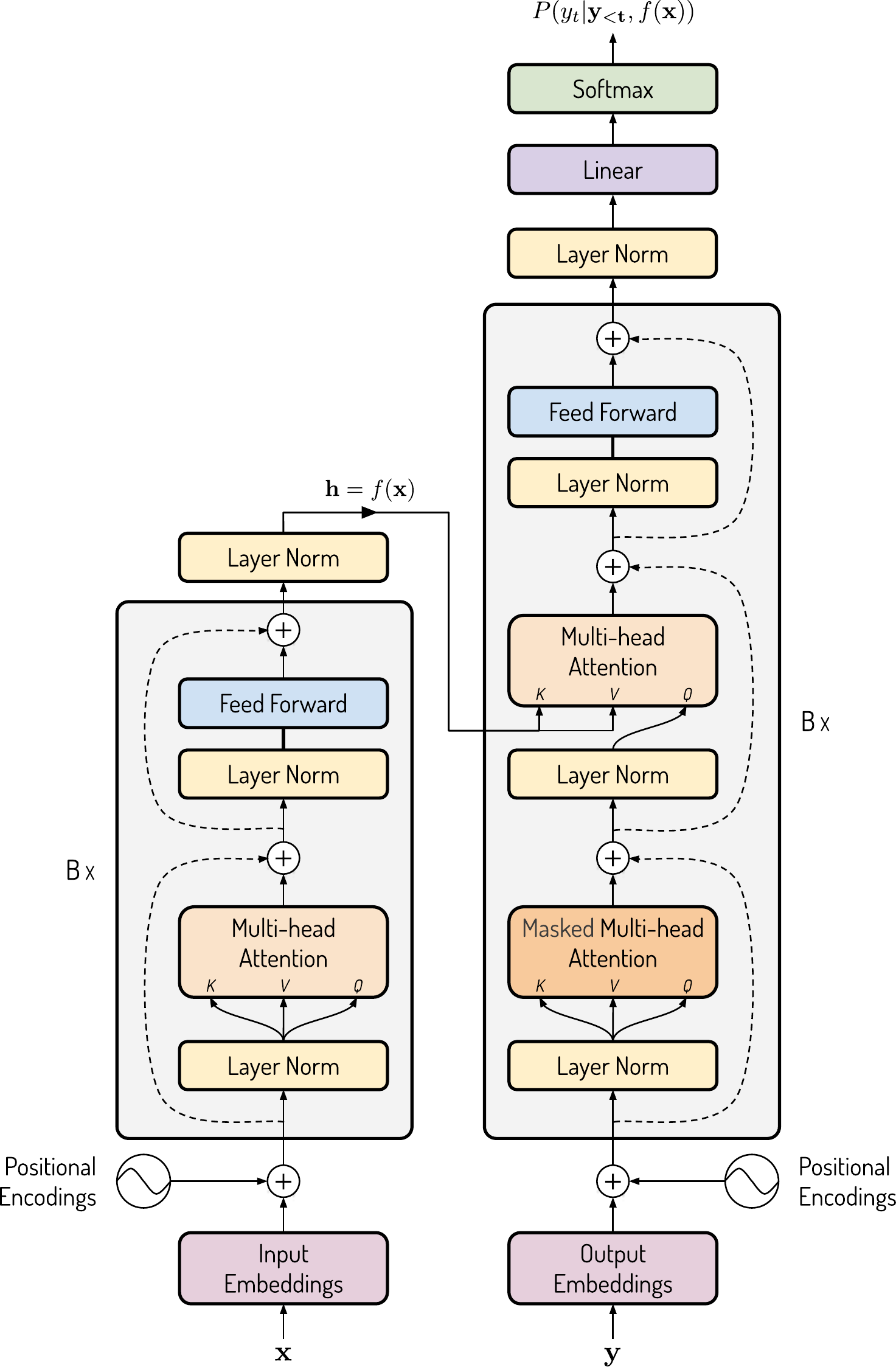}
\caption{The architecture of a Transformers NMT ~\shortcite{vaswani2017attention}: each block is repeated $B$ times to create a deep model. This is the more stable ``pre-norm'' variant~\shortcite{wang-etal-2019-learning-deep} where layer normalisation is applied prior to each sub-layer. The dashed lines denote the residual connections.}
\label{fig:tfnmt}
\end{figure}

Transformers NMT~\shortcite{vaswani2017attention} are the state-of-the-art sequence-to-sequence architectures equipped with deep encoder and decoder stacks that rely heavily on feed-forward layers in contrast to recurrent NMTs~\shortcite{sutskever2014sequence,Bahdanau2015NeuralMT}. Combined with the use of \I{self-attention} layers, these changes allow for (i) better gradient dynamics during training and thus deeper architectures, and (ii) different inductive biases than the left-to-right processing nature of recurrent NMTs. The overall diagram of a Transformers-based NMT is given in Figure~\ref{fig:tfnmt}. In the following, we briefly explain the encoder and the decoder blocks of Transformers.

\subsubsection{Encoder}
\label{sec:method_enc}
The transformer encoder $f()$ encodes the source sentence $\mathbf{x}$ into a latent representation $\mathbf{h} = f(\mathbf{x})$ using a series of operations. The basic block in the encoder transforms the input using a self-attention layer followed by a feed-forward layer. Layer normalisation~\shortcite{ba2016layernorm} and residual connections~\shortcite{he2016deep} are incorporated to achieve stability and improved training dynamics. This basic block (shown in the left part of Figure~\ref{fig:tfnmt}) is further replicated $B$ times in a vertical fashion so that each layer consumes the output of the previous one. The inputs to the first layer are the embeddings of the source sentence tokens $\mathbf{x} = [x_1, x_2, \dots, x_N]$, additively shifted by positional encodings. The latter is crucial to embed the positional information of words into the embeddings so that the representation is not invariant to the word order. The output of the last encoder layer is passed through a final layer normalisation. 

\subsubsection{Attention}
\label{sec:method_att}
\newcommand{\aq}{\mathbf{Q}}
\newcommand{\ak}{\mathbf{K}}
\newcommand{\av}{\mathbf{V}}
A key aspect of the Transformer architecture is the extensive use of several types of
attention layers. In its simplest form, the attention mechanism~\shortcite{Bahdanau2015NeuralMT} allows computing the weighted sum of a set of vectors $\av \in \mathbb{R}^{N\times D}$ where the weights are obtained based on dot product scores between keys $\ak \in \mathbb{R}^{N\times D}$ and queries $\aq \in \mathbb{R}^{N\times D}$. The scores are then normalised with the softmax operator to produce a valid probability distribution and finally multiplied with $\av$ as follows:
\begin{align}
    \text{A}(\aq, \ak, \av) = \text{softmax}\left(\frac{\aq \ak^T}{\sqrt{D}}\right) \av
    \label{eq:attn}
\end{align}

To further enrich the learned representations, multiple attention representations are computed in parallel, by supplying different projections of queries, keys and values as inputs to each attention function $\text{A}^{(i)}$ (Equation ~\ref{eq:atthead}). With $n$ attention heads, the final multi-head attention output is computed by projecting the concatenation of all attention head outputs as follows:
\begin{align}
    \text{A}^{(i)} &= \text{A}(\mathbf{W_{q}^{(i)}}\aq, \mathbf{W_{k}^{(i)}}\ak, \mathbf{W_{v}^{(i)}}\av)\label{eq:atthead} \\
    \text{MHA}(\aq, \ak, \av) &= \mathbf{W_{o}} \text{Concat}([\text{A}^{(1)}, \dots, \text{A}^{(n)}])
\end{align}

In the Transformer-based NMT, each encoder block includes a multi-head self-attention layer to capture the relationship between different source positions. Additionally, the decoder uses (i) \I{masked} multi-head self-attention to model the causal relationship between each position and the ones preceding it and (ii) cross-attention to integrate source semantics crucial to perform the translation. In fact, the only difference between self-attention and cross-attention is that the former sets $\aq = \ak = \av$ to the output of the previous layer whereas the latter uses the output of the \B{encoder} to set $\ak = \av = f(\mathbf{x})$ (Figure~\ref{fig:tfnmt}).

\subsubsection{Decoder}
\label{sec:method_dec}
Once the source sentence is encoded into the latent representation $\mathbf{h}$, the decoder sequentially generates the target sentence in an auto-regressive way. This means that the probability of the next target token ($P(y_t | \mathbf{y_{<t}}, \mathbf{h})$) is conditioned on the history ($\mathbf{y_{<t}}$) of target words predicted so far, in addition to the source sentence semantics encoded in $\mathbf{h}$. In this formulation, the whole decoder can be thought as a building block that implements the aforementioned probability term $P()$.

In terms of computation, a decoder block is very similar to an encoder one except that (i) the self-attention is now \I{masked} to enforce that the decoder is causal \ie it does not mistakenly look at future positions and (ii) a secondary multi-head attention known as \B{cross-attention}, integrates information from the \B{encoder} through the latent sentence representations $\mathbf{h}$ (\S~\ref{sec:method_att}). Finally, we train the model in an end-to-end way and minimise the negative log-likelihood of the sentence pairs in the training set $\mathcal{D}$:
\begin{equation}
\label{eq:condprob}
    \mathcal{L}_{\text{MT}} = - \sum_{i}^{|\mathcal{D}|} \log\left(P(\mathbf{y}^{(i)} | \mathbf{x}^{(i)})\right) \quad \text{where} \quad P(\mathbf{y} | \mathbf{x}) = \prod_{t=1}^{|\mathbf{y}|} P(y_t | \mathbf{y_{<t}}, \mathbf{h})
\end{equation}



\subsubsection{Our Baseline}
We use the \T{Base} Transformer~\shortcite{vaswani2017attention} configuration in all our experiments, where both the encoder and decoder have 6 layers ($B = 6$ in Figure~\ref{fig:tfnmt}), each attention layer has 8 heads, the model dimension is 512 and the feed forward layer size is 2048. Additionally, we share the parameters of the target and output language embedding matrix \shortcite{press2017using}. We should note that our implementation applies the ``pre-norm''~\shortcite{wang-etal-2019-learning-deep} formulation where the layer normalisation is placed prior to each sub-layer rather than after, to increase stability.

During training, we optimise the models using Adam \shortcite{kingma2014adam} and decay the learning rate with the \I{noam} scheduler~\shortcite{vaswani2017attention}. The initial learning rate, $\beta_1$ and $\beta_2$ are 0.2, 0.9 and 0.98, respectively. The learning rate is warmed up for 4,000 steps. We use a batch size of $32$, apply label smoothing with $\epsilon=0.1$~\shortcite{szegedy2016rethinking} and clip the gradients so that their norm is $1$~\shortcite{pascanu2014construct}. We train each system 3 times with different random seeds for a maximum of $100$ epochs, with early stopping based on the validation METEOR~\shortcite{meteor} score, which is the official metric used in all shared tasks in MMT \shortcite{barrault-etal-2018-findings}. The best checkpoint with respect to validation METEOR is selected to decode test set translations using the greedy search algorithm.

\subsection{Simultaneous NMT}
\label{sec:method_simt}
This section describes the different training strategies that are used in this work to realise simultaneous machine translation. Following the notation from \shortciteA{ma2019stacl}, we first define a function $g(t)$ that returns the number of source tokens read so far by the encoder, at a particular decoding timestep $t$. By definition, $0 \leq g(t) \leq |\mathbf{x}|$ for all values of $t$. We can formulate this by modifying Equation~\ref{eq:condprob} so that the source representation now depends on $g(t)$ as follows:
\begin{equation}
\label{eq:condprobwaitk}
    P(\mathbf{y} | \mathbf{x}) = \prod_{t=1}^{|\mathbf{y}|} P(y_t | \mathbf{y_{<t}}, f(\mathbf{x}_{\leq g(t)}))
\end{equation}
This generalisation allows us to define the conventional full sentence (consecutive) NMT as well, by using a constant function $g(t) = |\mathbf{x}|$ for all $t$ and all sentence pairs.


\begin{figure}[t!]
\centering
\includegraphics[width=0.75\textwidth]{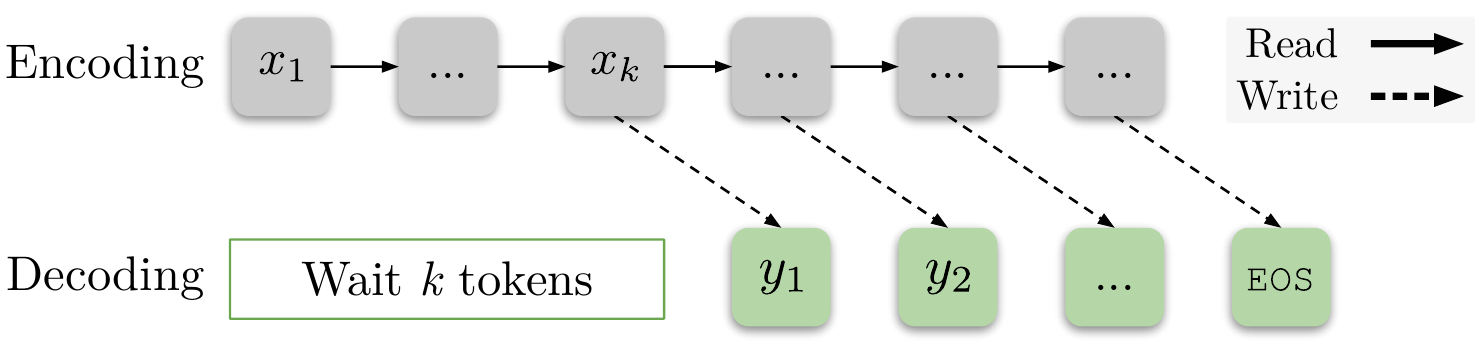}
\caption{\textsc{wait-k} decoding~\shortcite{ma2019stacl}: Initially, the decoder waits $k$ words to be read before committing the first translation. Afterwards, the algorithm switches back and forth between read and write actions until the end-of-sentence marker is generated by the decoder.}
\label{fig:waitk_decoding}
\end{figure}

\subsubsection{Wait-k Decoding}
The \textsc{wait-k} policy~\shortcite{ma2019stacl} simply amounts to selecting a particular function $g_{k}(t)$ that depends on a pre-determined $k$ value. This value determines the number of source tokens to be initially read by the encoder, before beginning the translation decoding process. Afterwards, the algorithm switches back and forth between read and write actions until the end-of-sentence marker (\T{EOS}) is predicted by the decoder. Mathematically speaking, the definition of $g_{k}(t)$ is as follows:
\begin{equation}
    \label{eq:gwaitk}
    g_{k}(t) = min(k + t - 1, |\mathbf{x}|)
\end{equation}
The \textsc{wait-k} decoding algorithm is a \I{fixed-delay} policy because the translator always lags behind the speaker by $k$ tokens. According to \shortciteA{ma2019stacl}, this policy is inspired by human interpreters who intuitively wait for some context to accumulate prior to starting the translation. For our multimodal purposes, 
having a fixed policy rather than an adaptive one is useful as fixed latency allows focused systematic analysis of quality improvements.
A depiction of the algorithm is given in Figure~\ref{fig:waitk_decoding}.

Traditionally, the encoder in the Transformer NMT is bi-directional due to the self-attention mechanism which allows each position to attend to each other one. In the simultaneous translation framework, however, this can be challenging as it implies that every time a new source token is read, the encoder’s representation $\mathbf{h}$ would need to be recomputed, leading to a quadratic runtime cost with respect to the source sentence length. Instead, we implement an \B{uni-directional} encoder following \shortciteA{elbayad2020efficient}, by employing a \B{masked} self-attention in the encoder. This prevents future positions that are not read so far from being attended to, similar to the self-attention layers in the decoder block.

\subsubsection{Simultaneous-aware Training}
\label{sec:method_simt_aware}
A simple way to perform simultaneous NMT is by first training a consecutive NMT model (\ie $g(t) = |\mathbf{x}|$) and then translating the test sets following the \textsc{wait-k} policy, where for each $k$ we would use a different choice of $g_k(t)$ at inference time. We refer to this approach as \textsc{wait-k decoding}. However, this creates a discrepancy between training and testing time, as the model will be exposed to partial source sentences at test time, although it was always trained on full sentences. For this reason, \shortciteA{ma2019stacl} also proposed to use the same $g_k(t)$ function at both training and inference time, an approach that we refer to as \textsc{wait-k training}. We follow this approach for our initial set of experiments.

\begin{algorithm}[t]
\small
\SetInd{0.5em}{0.5em}
\SetKwInOut{Input}{inputs}
\SetKwInOut{Output}{output}
\Input{The current mini-batch $B$}
\Output{The modified mini-batch $\hat B$}

$\hat B \leftarrow$ []\\

\For{$(\mathbf{x}, \mathbf{y})$ in $B$}{
    $c \sim \text{Uniform(0, 1)}$\I{\quad\quad\quad\quad// Apply truncation with $p=0.5$}\\
    \eIf{$c < 0.5$}
    {
      $l_x \sim \text{Uniform}(1, |\mathbf{x}|)$\I{\quad\quad// Randomly sample a source prefix length}\\
      $l_y \leftarrow \text{Round}\left(\dfrac{l_x . |\mathbf{y}|}{|\mathbf{x}|}\right)$\I{\quad\,// Keep the same proportion for the target}\\ 
      $l_y \leftarrow \text{max}(2, l_y)$\I{\quad\quad\quad\,\,\,// Always include the \T{BOS} token}\\
    }
    {
      $\mathbf{\hat x} \leftarrow \mathbf{x}$;
      $\mathbf{\hat y} \leftarrow \mathbf{y}$\I{\quad\quad\quad\,\,\,\,\,// No truncation}\\
    }
    $\hat B$.append($(\mathbf{\hat x}, \mathbf{\hat y})$)\\
}
\caption{Prefix training~\shortcite{niehues2018,arivazhagan2020re}}
\label{alg:prefix}
\end{algorithm}

Furthermore, we adopt a second simultaneous-aware training recipe called \B{prefix training}~\shortcite{niehues2018,arivazhagan2020re} which employ a simple data processing strategy to mitigate the aforementioned exposure bias between training and testing time. Specifically, for each sentence pair $(\mathbf{x}, \mathbf{y})$ in the mini-batch, we flip a fair coin to decide whether we will consider it for truncation or not. If it is considered, we first randomly sample a prefix length $l_x$ for the source sentence and truncate the corresponding target sentence with the same proportion as the source side (Algorithm~\ref{alg:prefix}). Unlike \textsc{wait-k training} which requires training a separate model for each value of $k$, we train a single \I{prefix} model and decode the final checkpoint using \textsc{wait-k decoding} across different values of $k$.

\subsection{Multimodal NMT}
In this section, we describe our take on integrating visual information to our consecutive and simultaneous NMT models.
For that, we reformulate the \T{encoder-attention} variant of~\shortciteA{caglayan2020simultaneous} for Transformer-based models. In what follows, we first describe the visual feature extraction pipeline and then present our multimodal NMT architecture in detail.

\subsubsection{Visual Features}
\label{sec:butd}
To represent visual semantics, we explore regional object-detection features
extracted using the popular \I{bottom up top down} (BUTD) approach~\shortcite{Anderson2017up-down}. BUTD combines the Faster R-CNN object detector~\shortcite{ren2015faster} with a ResNet-101 backend~\shortcite{he2016deep} to perform feature extraction. We use the provided model\footnote{https://hub.docker.com/r/airsplay/bottom-up-attention} which is pre-trained on the large-scale Visual Genome dataset~\shortcite{krishna2017visual}. Having 1,600 object labels in its inventory, the BUTD detector is quite rich and for that reason has been used in most previous work in cross-modal pre-training~\shortcite{lu2019vilbert,tan-bansal-2019-lxmert}.

For our purposes, we use the default settings of the detector and extract 36 regional features per each image. In other words, alongside the language representation $\mathbf{x} \in \mathbb{R}^{N\times D}$ of a given source sentence, the associated image is represented with a set of regional features $\mathbf{v} \in \mathbb{R}^{36\times 2048}$. The BUTD extractor is not further trained/fine-tuned during model training.

\begin{figure}[t!]
\centering
\includegraphics[width=0.9\textwidth]{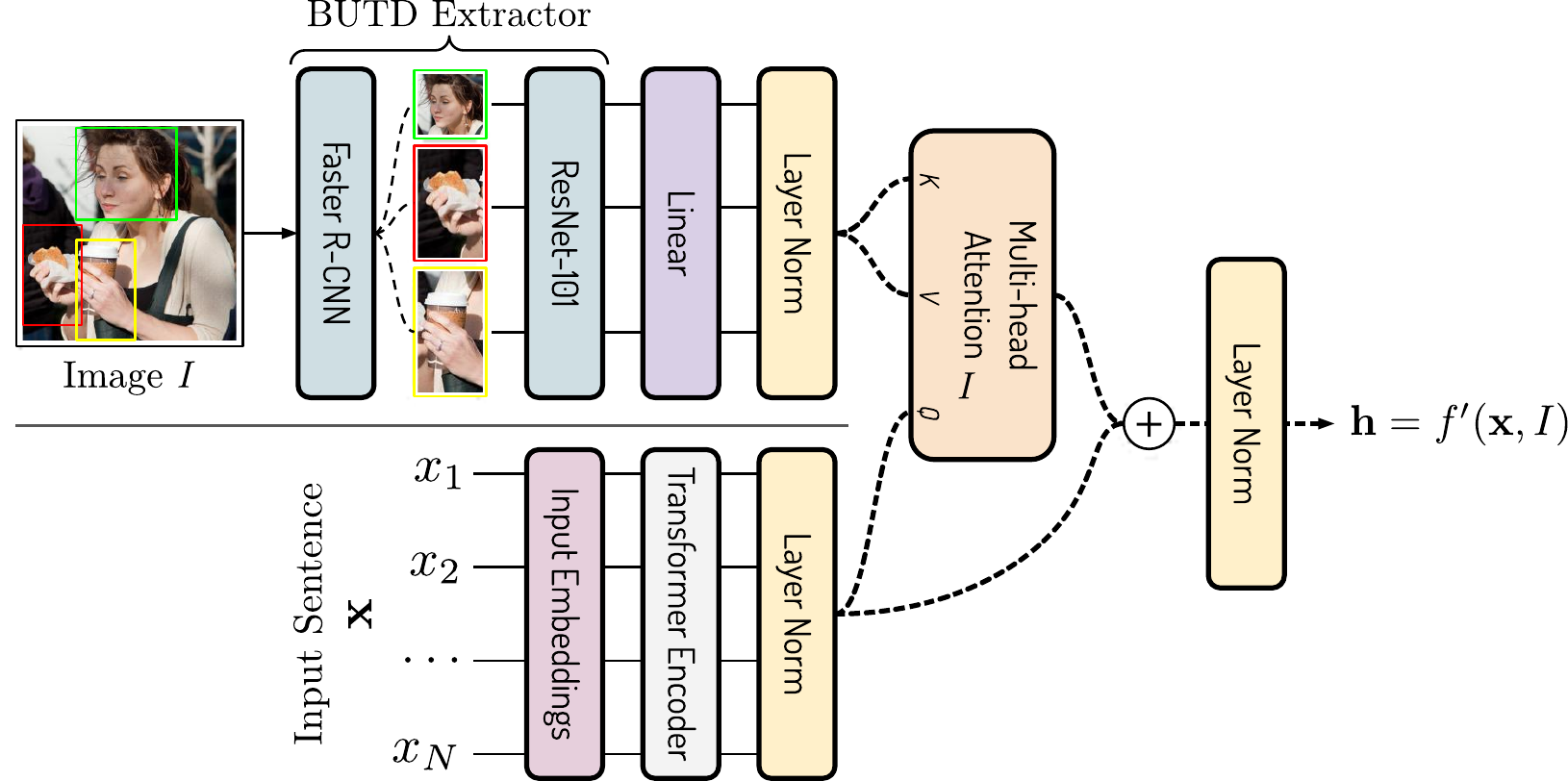}
\caption{Transformer-based multimodal \B{encoder} for simultaneous NMT: The upper stream encodes visual features, whereas the bottom one is the usual Transformer encoder for the input sentence. Dashed lines denote a collection of vectors.}
\label{fig:tfmmt}
\end{figure}

\subsubsection{Multimodality}
\label{sec:method_mmt}
Our multimodal MT approach reformulates the \T{encoder-attention} variant of~\shortciteA{caglayan2020simultaneous} for Transformer-based architectures. The main motivation for implementing an encoder-side cross-modal interaction is the nature of the simultaneous MT problem: since the source context is partial and grows gradually, the additional visual information can complement the missing context and allow the model to anticipate target words in a grounded way. Moreover, incorporating cross-modality at encoder side is also crucial for our second set of experiments regarding the supervision of the visual attention module, which aims at better language grounding (\S~\ref{sec:method_sup}).


Figure~\ref{fig:tfmmt} summarises the overall architecture of our MMT model, where the upper stream implements the visual representation module. This module (i) extracts the set of regional feature vectors $\mathbf{v} \in \mathbb{R}^{36\times 2048}$, (ii) projects them to the $D$-dimensional space yielding $\mathbf{v'} \in \mathbb{R}^{36\times D}$ and finally (iii) employs a multi-head\footnote{Throughout this work, we set the number of heads to 1 for the cross-modal interaction module. } attention layer for cross-modal interaction (CMI).
The key, value, query configuration of the cross-modal attention determines the nature of the interaction \ie we set $\ak = \av = \mathbf{v'}$ whereas the query $\aq$ receives the text encoder's output. This way, we get a cross-modal representation out of the attention layer which computes the weighted sum of regional feature projections ($\av$) based on the similarity between language representations ($\aq$) and regional features ($\ak$). Since this output is a linear combination of visual vectors only, we augment it with the text encoder's outputs using element-wise addition, similar to a residual connection. A final layer normalisation is applied on top to obtain the multimodal encoding $\mathbf{h} = f'(\mathbf{x}, I)$. The rest of the architecture is the same as Figure~\ref{fig:tfnmt} in the sense that $\mathbf{h}$ is passed to the cross-attention layer of the decoder. 

\subsubsection{Supervising the Visual Attention}
\label{sec:method_sup}
\begin{figure}[t!]
\centering
\includegraphics[width=0.96\textwidth]{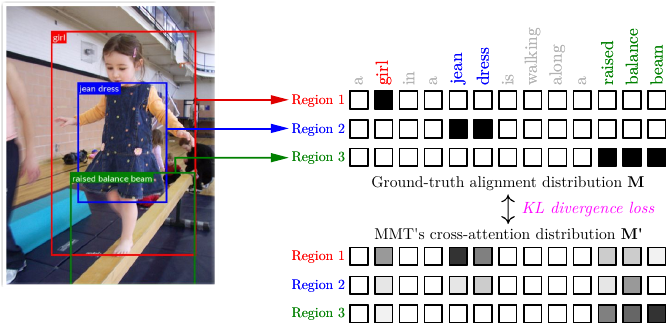}
\caption{Cross-modal supervision of attention: The attention distribution from the model is pulled towards the ground-truth alignment matrix obtained from the Flickr30k entities dataset, using KL-divergence. Greyed out token positions do not contribute to the loss.}
\label{fig:supervision}
\end{figure}

The cross-modal relationship between language and visual representations in attentive MMT models is generally learned in an unsupervised way, as the MMT model is only trained through a sequence-to-sequence cross-entropy objective. This is also the approach that we followed for our MMT model in \S~\ref{sec:method_mmt}.

In this section, we are interested in whether explicit supervision of the visual attention could  introduce an inductive bias so that the anticipation ability -- and therefore the translation quality -- of our simultaneous MMT models is further improved. We devise a \B{multi-task learning} scheme where in addition to the cross-entropy objective used for the MT task (Equation~\ref{eq:condprob}), we employ an auxiliary loss which tries to bring together the cross-modal attention distribution $\mathbf{M'}$ computed by the CMI module and the ground-truth alignment $\mathbf{M}$. An illustration of the predicted attention and its ground-truth distribution is provided in Figure~\ref{fig:supervision}.

\paragraph{Ground-truth alignments.}
In order to supervise the visual attention during training, we require a labeled dataset of phrase-region annotations for the Multi30k dataset. Since the Multi30k dataset is derived from the Flickr30k image captioning dataset, we rely on the Flickr30k Entities dataset~\shortcite{plummer-etal-2015-collecting} which provides human-annotated bounding boxes for noun phrases in the English image captions. For instance in Figure~\ref{fig:supervision}, we can see that each of the phrases ``\T{a girl}'', ``\T{a jean dress}'' and ``\T{a raised balance beam}'' is mapped to a bounding box\footnote{The number of bounding boxes for a given noun phrase is not limited to one.} that denotes the object referred.

Since the ground-truth regions are different from the regions predicted by the pre-trained BUTD detector (\S~\ref{sec:butd}), we \B{re-extract} regional features using the same pre-trained BUTD detector but from the set of regions provided by the Entities dataset. On average,
we end up with 4.3 bounding box annotations per image, which is substantially lower compared to the fixed number of 36 regions that we extracted from the pre-trained BUTD detector. Finally, we should note that the Entities dataset does not provide any annotations for the \T{test2017} and \T{testCOCO} test splits of Multi30k dataset.

%


\paragraph{Training.}
For a given source sentence $\mathbf{x}$ with $N$ words, first an alignment matrix $\mathbf{M} \in \mathbb{R}^{R\times N}$ is formed where $R$ denotes the total number of ground-truth bounding box annotations for the overall sentence. We set $\mathbf{M}_{[i,j]} = 1$ if the word $x_j$ is associated with the region $i$. If $k > 1$ region associations exist for the word $x_j$, the probability mass assigned to each region becomes $1/k$. The columns of the matrix that refer to words without any bounding box associations are not taken into account when computing the final alignment loss.

\paragraph{Alignment loss.} The Kullback-Leibler (KL) divergence is a measure of how much a given probability distribution $Q$ differs from a reference distribution $P$. In other words, minimising the KL-divergence between the predicted cross-modal attention distribution ($Q$) and the ground-truth alignments from the entities dataset ($P$) allows the model to generate a visual attention distribution that is closer to the human-labeled annotations. The final multi-task objective combines the MT loss and the alignment loss with coefficients $\alpha = \beta = 1$:
\begin{align}
    \mathcal{L} &= \alpha \mathcal{L}_{\text{MT}} + \beta  D_{\text{KL}}\left(P=\mathbf{M}\,\,||\,\,Q=\mathbf{M'}\right)
\end{align}

\paragraph{Fine-tuned supervision.} In addition to training the supervised MMT from scratch, we also experiment with taking the best checkpoints of the MMT models with unsupervised visual attention ($\alpha = 1$, $\beta = 0$) and fine-tuning them with the alignment loss enabled ($\alpha = \beta = 1$).
For these particular variants, we disable the learning rate scheduling and lower the learning rate to $1e-5$. We track performance with respect to validation set METEOR and keep the best checkpoint for further decoding of the test sets.

\subsection{Dataset}
\label{sec:dataset}
We use the Multi30k dataset~\shortcite{elliott-etal-2016-multi30k}\footnote{https://github.com/multi30k/dataset}, which is a multi-lingual extension to the Flickr30k image captioning dataset~\shortcite{young-etal-2014-image}. Specifically, Multi30k provides a training set of 29,000 examples where each example corresponds to an image $I$ and its English caption $\mathbf{x}$ from Flickr30k, extended with the German translation $\mathbf{y}$ of the English caption. In other words, both training and testing samples are triplets in the form of $\{I, \mathbf{x}, \mathbf{y}\}$.
The dataset is later extended with French~\shortcite{elliott-etal-2017-findings} and Czech~\shortcite{barrault-etal-2018-findings} translations, making it the standard dataset for work on MMT, simultaneous MMT~\shortcite{caglayan2020simultaneous,imankulova-etal-2020-towards}, as well as other multilingual multimodal tasks.

\begin{table}[t!]
\centering
\renewcommand{\arraystretch}{1.2}
\resizebox{.85\textwidth}{!}{%
\begin{tabular}{l|rc|rc|rc|rc|r}
\hline
              & \MC{2}{c|}{English}     & \MC{2}{c|}{German}  & \MC{2}{c|}{French}  & \MC{2}{c|}{Czech}  &        \\
  Split       &    Words & Len   & Words & Len  & Words & Len  & Words & Len &  \# Sents \\ \hline
  \T{train}   & 380K     & 13.1        & 364K     & 12.6    & 416K     & 14.4    & 298K  & 10.3      & 29,000 \\
  \T{val}     & 13.4K    & 13.2        & 13.1K    & 12.9    & 14.6K    & 14.4    & 10.4K & 10.2      & 1,014  \\ \hline
  \T{test2016}& 13.0K    & 13.1        & 12.2K    & 12.2    & 14.2K    & 14.2    & 10.5K & 10.5      & 1,000  \\
  \T{test2017}& 11.4K    & 11.4        & 10.9K    & 10.9    & 12.8K    & 12.8    & -     & -         & 1,000  \\
  \T{testCOCO}&  5.2K    & 11.4        &  5.2K    & 11.2    &  5.8K    & 12.5    & -     & -         & 461    \\
\hline
\end{tabular}}
\caption{Multi30k dataset statistics: ``Words'' denote the total number of words in a split whereas ``Len'' is the average number of words per sentence in that split.}
\label{tbl:multi30k}
\end{table}

We experiment with all three language pairs, namely, English\ra German, English\ra Czech and English\ra French. The training, validation and the \T{test2016} test sets are available for all language directions, whereas the test sets \T{test2017} and \T{testCOCO} sets are only available for German and French. The latter test set is specifically designed~\shortcite{elliott-etal-2017-findings} to contain at least one ambiguous word per image caption where images are selected from the COCO dataset~\shortcite{chen-etal-2015-coco} instead of the in-domain Flickr30k dataset. Some common statistics of the dataset are provided in Table~\ref{tbl:multi30k}.

\paragraph{Preprocessing.} We use Moses tools~\shortcite{koehn-etal-2007-moses} to lowercase, punctuation-normalise and tokenise the sentences with \I{hyphen splitting} option enabled. We then create word vocabularies using the \I{training} subset of each language.
The resulting English, French, German and Czech vocabularies have 9.8K, 11K, 18K and 22.2K words, respectively.
We do not use sub-word segmentation approaches to avoid their potential side effects for simultaneous MT and to be able to analyse the grounding capability of the models more easily when it comes to cross-modal attention. We note that word-level MT performs as well as sub-word segmentation on this particular dataset, according to~\shortcite{caglayan-thesis-2019}.

\section{Results} 
\label{sec:results}
\subsection{Unimodal Simultaneous MT}
Our first set of experiments focus on how different training regimes impact the translation quality for Transformer-based unimodal (i.e. text-only) simultaneous MT. For this, we compute METEOR scores of English-Czech, English-German and English-French MT systems, across three different test sets of Multi30k. Table~\ref{tbl:unimodal_all} summarises the results obtained by performing \textsc{wait-k} decoding policy with $k\in \{1,2,3\}$, across the methods described in \S~\ref{sec:method_simt}.

\begin{table}[t!]
\centering
\def\arraystretch{1.2}
\resizebox{.85\textwidth}{!}{%
\begin{tabular}{ccccc|ccc|ccc}
 \hline
 & & \MC{3}{c}{2016}  & \MC{3}{c}{2017} & \MC{3}{c}{COCO} \\
 & &
 \textsc{Cons} & \textsc{Tr} & \textsc{Pref} &
 \textsc{Cons} & \textsc{Tr} & \textsc{Pref} &
 \textsc{Cons} & \textsc{Tr} & \textsc{Pref} \\
 \hline
\parbox[t]{2mm}{\multirow{3}{*}{\rotatebox[origin=c]{90}{\B{$k=1$}}}} 
 & \textsc{En-Cs} & \Tt{17.1} & 13.0      & \Tf{22.4} & - & - & - & - & - & - \\
 & \textsc{En-De} & 40.3      & \Tt{41.3} & \Tf{48.8} & \Tt{37.2} & 35.8 & \Tf{43.2} & \Tt{35.9} & 33.3 & \Tf{39.6} \\
 & \textsc{En-Fr} & \Tt{59.5} & 58.7      & \Tf{63.0} & \Tt{55.6} & 52.6 & \Tf{58.0} & \Tt{52.6} & 48.9 & \Tf{53.7} \\
 \hline

\parbox[t]{2mm}{\multirow{3}{*}{\rotatebox[origin=c]{90}{\B{$k=2$}}}}
 & \textsc{En-Cs} & \Tt{21.2} & 19.0      & \Tf{24.1} & - & - & - & - & - & - \\
 & \textsc{En-De} & 46.7      & \Tt{48.7} & \Tf{52.4} & 42.4      & 42.4 & \Tf{46.3} & \Tt{41.3} & 39.4 & \Tf{42.4} \\
 & \textsc{En-Fr} & \Tt{66.9} & 66.3      & \Tf{68.1} & \Tt{62.0} & 60.7 & \Tf{62.5} & \Tf{58.7} & 56.7 & \Tt{58.4} \\
 \hline

\parbox[t]{2mm}{\multirow{3}{*}{\rotatebox[origin=c]{90}{\B{$k=3$}}}}
 & \textsc{En-Cs} & \Tt{23.9} & 23.3      & \Tf{26.5} & - & - & - & - & - & - \\
 & \textsc{En-De} & 51.4      & \Tt{51.8} & \Tf{53.9} & \Tt{46.0} & 45.1 & \Tf{47.8} & \Tf{44.4} & 41.4 & \Tt{43.6} \\
 & \textsc{En-Fr} & \Tt{71.4} & 71.3      & \Tf{71.5} & \Tf{65.9} & 65.4 & \Tt{65.2}& \Tf{61.1} & 59.8 & \Tt{60.1}  \\
\hline
\end{tabular}}
\caption{METEOR comparison of consecutive training (\textsc{Cons}), \textsc{wait-k} training (\textsc{Tr}) and prefix training (\textsc{Pref}) approaches for \B{unimodal} simultaneous MT. All systems are decoded with the \textsc{wait-k} policy where $k \in \{1,2,3\}$. The scores are averages of three runs with different seeds. Best systems are indicated in \B{bold} typeface.}
\label{tbl:unimodal_all}
\end{table}

First, we observe that the prefix training method yields substantially higher METEOR scores than the other two approaches in general. 
\textsc{wait-k} training achieves the lowest quality across all language pairs and test sets, an observation in line with the previous findings of \shortciteA{caglayan2020simultaneous}. Overall, the quantitative results corroborate our initial hypothesis regarding the exposure bias between training and test time (\S~\ref{sec:method_simt_aware}) and show that the prefix training is indeed a good choice in-between the full-sentence (consecutive) training and the aggressive \textsc{wait-k} training approaches.

For \B{English-Czech} particularly, we observe that when $k=1$, \textsc{wait-k} training is considerably worse ($\Downarrow$4.1 \textsc{Meteor}) than the other approaches. We hypothesise that the lower scores for the English-Czech pair in general is likely to be caused by the fact that, unlike English, German and French, Czech has no articles preceding a noun. This means that when generating the first word of the Czech translation, the probability distribution will always generate the same word, which is likely to be the most common target sentence prefix in the training set. This is illustrated in Table~\ref{tbl:unimodal_ex} where the Czech model translates ``woman'' as ``Muž'' (man) whereas the German and French models are able to read the source word ``woman'' just before translating it\footnote{Although the same issue also affects the article choices for gender-marked languages such as French and German, the arbitrary choices are more likely to match the correct article in the target language, since the number of possible translations for articles is much smaller.}. We believe that the incorporation of the visual modality would allow simultaneous MT models to handle such cases in a better way.

Finally, although the lack of context affects all three methods at decoding time, the reason why \textsc{wait-k} training lags dramatically behind the other two for Czech is related to the fact that, \textsc{wait-k} training is ineffective in terms of leveraging the training resources, especially when the vocabulary is sparse. This makes Czech the most challenging pair across the pairs explored as its vocabulary has 22K unique tokens compared to 11K for French.


\begin{table}[t!]
\centering
\resizebox{.7\textwidth}{!}{%
\begin{tabular}{rl}
\toprule
\T{SRC:} & A woman \G{holding a bowl of food in a kitchen.} \\
\midrule
\T{CS:} & \W{Muž} v \G{červeném triku drží své jídlo.} \\
\T{DE:} & \W{Ein} \C{Frau} \G{hält eine schüssel mit Essen in einer Küche.} \\
\T{FR:} & \W{Un} \C{femme} \G{tenant un bol de nourriture dans une cuisine.}\\
\bottomrule
\end{tabular}}
\caption{Examples showing the effect of the lack of context for the decoder early in the sentence when using \textsc{wait-1} decoding: greyed out words are those that have not been read or written yet. \W{Red} and \C{blue} indicate \W{incorrect} and \C{correct} translations, respectively.}
\label{tbl:unimodal_ex}
\end{table}


\subsection{Multimodal Simultaneous MT}
Based on the more promising results of the prefix training regime for the unimodal MT experiments, from now on we focus on prefix training across all  multimodal experiments.
We begin by analysing the impact of incorporating visual information into our Transformer-based simultaneous NMT models, following the architecture described in \S~\ref{sec:method_mmt}. Table~\ref{tbl:mmt_all} compares METEOR scores between the baseline unimodal NMT and the MMT models across all language pairs and test sets.

\begin{table}[t!]
\centering
\def\arraystretch{1.2}
\resizebox{.85\textwidth}{!}{%
\begin{tabular}{cl|c|ccc|ccc}
\hline
 & & \MC{1}{c|}{\textsc{En-Cs}} & \MC{3}{c|}{\textsc{En-De}} & \MC{3}{c}{\textsc{En-Fr}} \\
 &
 & 2016
 & 2016 & 2017 & CC
 & 2016 & 2017 & CC \\
\hline
\parbox[t]{2mm}{\multirow{2}{*}{\rotatebox[origin=c]{90}{$k=1$}}}
& \nmt  &\Tf{22.4} & \Tf{48.8}     & 43.2     & 39.6        & 63.0     & 58.0     & 53.7        \\
& \mmt  & 21.9     & \Tf{48.8}     &\Tf{43.3} &\Tf{39.7}    &\Tf{64.5} &\Tf{58.2} &\Tf{54.2}    \\
\hline
\parbox[t]{2mm}{\multirow{2}{*}{\rotatebox[origin=c]{90}{$k=2$}}}
& \nmt  & 24.1     &\Tf{52.4} & 46.3     & 42.4        & 68.1     & 62.5     & 58.4        \\
& \mmt  &\Tf{24.4} & 52.1     &\Tf{46.4} &\Tf{43.1}    &\Tf{69.0} & \Tf{63.2}& \Tf{58.6}   \\ 
\hline
\parbox[t]{2mm}{\multirow{2}{*}{\rotatebox[origin=c]{90}{$k=3$}}}
& \nmt  & 26.5     &\Tf{53.9} & \Tf{47.8}     & 43.6        & 71.5     & 65.2     & 60.1 \\
& \mmt  & 26.5     & 53.5     & \Tf{47.8}     &\Tf{44.1}    &\Tf{71.8} &\Tf{65.7} &\Tf{60.4} \\ 

\hline
\end{tabular}}
\caption{METEOR comparison of \B{unimodal} and \B{multimodal} simultaneous MT under the \B{prefix training} framework: the scores are averages of three runs with different seeds. Best systems are indicated in \B{bold} typeface.}
\label{tbl:mmt_all}
\end{table}

Overall, our encoder-based cross-modal interaction method improves METEOR scores up to +1.5 points (\textsc{En-Fr 2016}) when compared to unimodal NMT models. This shows that the addition of the visual modality is beneficial for simultaneous MT, especially in low latency scenarios. In particular, consistent quality improvements are observed on the \T{testCOCO} test split, which (i) does not come from the Flickr30k distribution and (ii) contains more ambiguous words by construction. 

For certain language pairs such as English\ra Czech and English\ra German, the additional modality brings little to no improvements, especially on the \T{test2016} test set. However, we will observe different trends once we add supervision (\S~\ref{sec:sup_att}).
Finally, the \B{English}\ra\B{French} language pair benefits the most from the visual modality on average, in terms of METEOR scores. We find it interesting that as the lexical diversity of the target language (\ie sparsity of the training set vocabulary) increases, the improvements due to multi-modality decrease. This hints at the fact that the improvements can be more pronounced if the models are (pre-)trained on larger linguistic resources, an exploration that we leave to future work.

\subsubsection{Supervised Attention}
\label{sec:sup_att}
Our next set of experiments aim to understand whether guiding the alignment between the visual features and source words during training helps improve the translation and grounding ability of the MMT models. We carry out two different experiments: first, we train a multimodal simultaneous translation model from \textsc{Scratch}, using both the translation and attention losses. As an alternative, we also experiment with a \textsc{Fine-Tuned} version of the best (unsupervised visual attention) MMT checkpoints with a reduced learning rate (\S\ref{sec:method_sup}).

The METEOR scores for supervision experiments are reported in Table~\ref{tbl:mmt_supervised}. Overall, the scores highlight that (i) both variants of supervision are beneficial to the MMT models, which now show more consistent improvements over both the baselines and the MMT models with unsupervised visual attention and (ii) the \textsc{Fine-Tuned} model produces consistently higher scores, with improvements of up to $1.9$ METEOR points (\textsc{En-Fr 2016}) with respect to the unimodal baseline.

\begin{table}[t!]
\centering
\def\arraystretch{1.2}
\resizebox{.85\textwidth}{!}{%
\begin{tabular}{=c +l | +c | +c +c +c| +c +c +c}
\hline
 & & \MC{1}{c|}{\textsc{En-Cs}} & \MC{3}{c|}{\textsc{En-De}} & \MC{3}{c}{\textsc{En-Fr}} \\
 &
 & 2016 
 & 2016 & 2017 & CC
 & 2016 & 2017 & CC \\
\hline
\parbox[t]{2mm}{\multirow{4}{*}{\rotatebox[origin=c]{90}{$k=1$}}}
& \nmt     &22.4     &48.8     &43.2     &39.6     &63.0     &58.0     &53.7      \\
& \mmt     &21.9     &48.8     &43.3     &39.7     &64.5     &58.2     &54.2      \\
& \mmtsup  &21.8     &48.8     &43.1     &39.3     &64.3     &58.6     &53.8      \\
& \mmtsupf &\Tf{22.5}&\Tf{49.2}&\Tf{43.5}&\Tf{39.9}&\Tf{64.9}&\Tf{58.9}&\Tf{54.8} \\
\hline

\parbox[t]{2mm}{\multirow{4}{*}{\rotatebox[origin=c]{90}{$k=2$}}}
& \nmt     &24.1 &52.4     &46.3     &42.4     &68.1     &62.5     &58.4          \\
& \mmt     &\Tf{24.4} &52.1     &46.4     &\Tf{43.1}&69.0     &63.2     &58.6          \\ 
& \mmtsup  &\Tf{24.4} &52.3     &46.1     &42.3     &69.2     &63.1     &57.9          \\
& \mmtsupf &\Tf{24.4} &\Tf{52.9}&\Tf{46.8}&43.0&\Tf{69.5}&\Tf{63.9}&\Tf{58.9}     \\
\hline

\parbox[t]{2mm}{\multirow{4}{*}{\rotatebox[origin=c]{90}{$k=3$}}}
& \nmt     &26.5     &53.9     &47.8     &43.6     &71.5     &65.2     &60.1      \\
& \mmt     &26.5  &53.5     &47.8     &\Tf{44.1}&71.8     &65.7     &60.4      \\ 
& \mmtsup  &26.6     &53.7     &47.7     &43.1     &71.9     &65.4     &59.7      \\
& \mmtsupf &\Tf{27.0}&\Tf{54.3}&\Tf{48.4}&\Tt{43.8}&\Tf{72.5}&\Tf{66.4}&\Tf{60.5} \\
\hline
\end{tabular}}
\caption{The impact of attention supervision with respect to baseline NMT and MMT systems: METEOR scores are averages of three runs with different seeds. Best systems are indicated in \B{bold} typeface.}
\label{tbl:mmt_supervised}
\end{table}

Interestingly, the \textsc{Fine-Tuned} variant always outperforms the \textsc{Scratch} models. One possible explanation for this is that training the MMT model from scratch with both objectives might bias the encoder towards producing representations that align well with image regions, but not necessarily optimal for the MT decoder. To get a better insight on this, we performed a small study where we repeated the supervision experiments by lowering the alignment loss weight from $\beta=1$ to $\beta \in \{0.1, 0.5\}$. The results showed that although there are cases where tuning the coefficient yields improved scores for some language pairs, the scores are not as consistent as $\beta=1$ across different \textsc{wait-k} decodings and test sets.
We leave further exploration of this to future work.

\subsubsection{Prefix Translation Accuracy}
Automatic evaluation metrics are often not very suitable for tasks such as image captioning or MMT, as they can fail to capture non-systematic \& minor changes~\shortcite{caglayan-etal-2020-curious}. This issue is even more pronounced in simultaneous translation, where models are more likely to make mistakes early in the translation (due to limited source context), which could have a negative impact in the overall translation quality. In this section, we provide an analysis focused on the translation accuracy of source prefixes. Specifically, 
we count the number of matches between the first $n$ words of each translation and its reference sentence, and divide it by the number of test set sentences.

\begin{figure}[pt]
\centering
\includegraphics[width=0.99\textwidth]{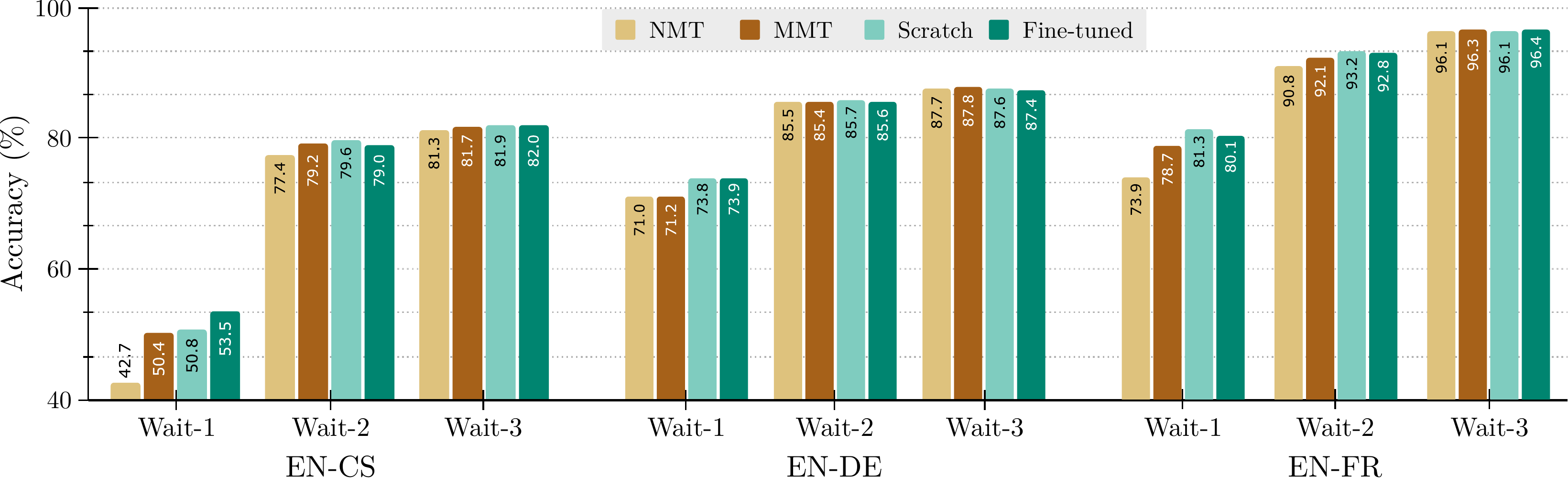}
\caption{Unigram prefix accuracy of simultaneous \textsc{wait-k} variants: the numbers are obtained from the \T{test2016} test set, averaged across three runs.}
\label{fig:prefix_acc}
\end{figure}

Figure~\ref{fig:prefix_acc} shows the unigram prefix accuracy of each \textsc{wait-k} decoded simultaneous MT model, across the three language pairs explored. The results reveal several things: (i) the supervision is globally beneficial, although its contribution diminishes as $k$ increases, and (ii) although not reflected by the METEOR scores (Table~\ref{tbl:mmt_supervised}), the incorporation of the visual modality (and further supervision) boosts the accuracies for \B{English\ra Czech} substantially (especially for \textsc{wait-1}). For the latter, this amounts to a 25\% relative improvement by the \textsc{Fine-tuned} supervision with respect to the baseline NMT. Similar trends were observed for bigram and trigram accuracy for \B{English\ra Czech} and \B{English\ra French}. 

\subsubsection{Comparison to State-of-the-art}
Finally, we conclude our quantitative analyses by comparing the performance of our best MMT models to state-of-the-art in simultaneous multimodal MT. Table~\ref{tbl:mmt_sota} provides a summary of the results in terms METEOR and also BLEU~\shortcite{papineni2002bleu}, since some previous work only reports the latter.
Recall that (i) \shortciteA{imankulova-etal-2020-towards} and \shortciteA{caglayan2020simultaneous} rely on recurrent models rather than Transformers, (ii) use different visual features and (iii) train their models using either consecutive or \textsc{wait-k} approaches rather than our best \I{prefix training} methodology.

According to Table~\ref{tbl:mmt_sota}, the scores clearly show that when combined with the \textsc{Fine-Tuned} attention supervision variant, our Transformer-based MMT models achieve state-of-the-art BLEU and METEOR scores across the board, with the exception of English\ra German with larger $k$. For this language direction, our \textsc{wait-2} and \textsc{wait-3} systems slightly lag behind \shortciteA{caglayan2020simultaneous} in terms of BLEU. This is probably due to our reliance on METEOR for early-stopping and best checkpoint selection rather than the perplexity, which was used by \shortciteA{caglayan2020simultaneous}.

\begin{table}[t]
\centering
\def\arraystretch{1.2}
\resizebox{.85\textwidth}{!}{%
\begin{tabular}{cl|cc|cc|cc}
\hline
 & & \MC{2}{c|}{\textsc{En-Cs}} & \MC{2}{c|}{\textsc{En-De}} & \MC{2}{c}{\textsc{En-Fr}} \\
 &
 & BL & MT 
 & BL & MT 
 & BL & MT  \\
\hline
\parbox[t]{2mm}{\multirow{4}{*}{\rotatebox[origin=c]{90}{$k=1$}}}
& \shortciteA{imankulova-etal-2020-towards}   & 9.1  & -  & 19.9   & - &   32.5   &  -  \\
& \shortciteA{caglayan2020simultaneous} \textsc{(Dec Att)}  &  -  & - & 21.2  & 45.9 &  42.1    & 64.5 \\
& \shortciteA{caglayan2020simultaneous} \textsc{(Enc Att)}  &  -  & -  & 21.3 & 44.7 &  41.7  & 64.6 \\
& \textsc{Our Best} &  \B{12.5}  & \B{22.5}&  \B{21.4}  & \B{49.4} &  \B{42.9} & \B{64.9}  \\
\hline

\parbox[t]{2mm}{\multirow{4}{*}{\rotatebox[origin=c]{90}{$k=2$}}}
& \shortciteA{imankulova-etal-2020-towards}     &  -  &  -  & -  & - & - & - \\
& \shortciteA{caglayan2020simultaneous} \textsc{(Dec Att)} &  -  & - &  \B{28.1}  & 50.4 &  49.2   & 68.7 \\
& \shortciteA{caglayan2020simultaneous} \textsc{(Enc Att)}  &  -  & - &  27.6  & 49.7  &  49.2  & 69.1  \\
& \textsc{Our Best} &  \B{16.1}  & \B{24.4}  & 28.0   & \B{52.9}  &   \B{51.5}   &  \B{69.5}  \\
\hline

\parbox[t]{2mm}{\multirow{4}{*}{\rotatebox[origin=c]{90}{$k=3$}}}
& \shortciteA{imankulova-etal-2020-towards}     &  19.4 & -  & 28.8   & -    & 44.0    &  -    \\
& \shortciteA{caglayan2020simultaneous} \textsc{(Dec Att)}  & -  & -  &  \B{32.2}  & 53.7 &  54.6    &  71.8 \\
& \shortciteA{caglayan2020simultaneous} \textsc{(Enc Att)}  & -  & -  & 31.4   &  53.1 & 54.8   &   72.0 \\
& \textsc{Our Best} & \B{21.8}  & \B{27.0} & 31.3 & \B{54.3}  & \B{56.2}    &  \B{72.5}  \\
\hline
\end{tabular}}
\caption{Comparison of our best MMT system (fine-tuned supervised attention) with other state-of-the-art simultaneous MMT models on the \T{test2016} set. Best systems are indicated in \B{bold} typeface. BL and MT denote BLEU and METEOR, respectively.}
\label{tbl:mmt_sota}
\end{table}

\subsection{Qualitative Insights}
In this section, we conduct further analysis focusing on qualitative aspects of the trained MMT models in terms of visual understanding.

\subsubsection{Grounding Ability}
We begin our analyses by measuring the grounding ability of our MMT systems on the \T{test2016} test set, for which we have ground-truth region-phrase annotations from Flickr30k dataset (\S~\ref{sec:dataset}). More formally, we obtain the cross-modal attention weights from our models during the translation decoding of the test set sentences.
Next, similar to \shortciteA{rohrbach2016grounding} and \shortciteA{wang-specia-2019}, we compute the intersection over union (IoU) between the most-attended region and the ground-truth bounding box. If multiple ground-truth boxes exist for a given phrase, we take the ground-truth box that yields the maximum IoU score.
Table~\ref{tbl:mmt_grounding} provides a summary of the results across the language pairs explored. Unsurprisingly, we notice that the MMT models with unsupervised visual attention obtain the lowest grounding accuracy across the board. The supervision from \textsc{Scratch} achieves the highest accuracy, surpassing the models with unsupervised visual attention by $18\%$ on average.
The \textsc{Fine-Tuned} supervision approach seems also quite helpful in terms of grounding, with accuracies slightly lagging (around $2\%$) behind the \textsc{Scratch} systems. Another interesting observation is that the grounding ability does not seem to depend on the choice of target language, \eg all \textsc{Fine-Tuned} MMT models achieve around $89\%$ accuracy.
This is to be expected, since during training, the source sentences and the provided regions never change across different language directions.

\subsubsection{Performance on Butd Regions}
The previous experiment relied on regional features that were extracted from the ground-truth bounding boxes at training, inference and evaluation.
To understand whether the grounding ability generalises across features extracted from \B{non-gold} regions, we now switch to the 36 region proposals that we extracted from the pre-trained BUTD object detector (\S~\ref{sec:butd}).
Instead of working in the visual coordinate space, which may be noisy due to the fine-grained nature of bounding box annotations, given that BUTD also provides a wider range of object categories, we also measure the grounding ability in the \B{linguistic space}. Specifically, instead of the coordinates of the BUTD and ground-truth regions, we use the predicted object label $\hat y$ (from the BUTD detector) and the ground-truth noun annotation $y$, respectively. We then compute exact match accuracy (\textsc{Em}) and cosine similarity\footnote{For the cosine similarity approach, we ensure maximum correspondence by lemmatising both words using the \T{NLTK} toolkit~\shortcite{bird-loper-2004-nltk}.} (\textsc{Cos}) between the lowercased versions of these words, by using pre-trained GloVe~\shortcite{pennington-etal-2014-glove} word embeddings.
\begin{table}[t]
\centering
\def\arraystretch{1.2}
\resizebox{.8\textwidth}{!}{%
\begin{tabular}{l|c|c|c}
\hline
 & \MC{1}{c|}{\textsc{En-Cs}} & \MC{1}{c|}{\textsc{En-De}} & \MC{1}{c}{\textsc{En-Fr}} \\
\hline

\textsc{Mmt} & 73.8\% ($\pm$ 0.7) & 72.1\% ($\pm$ 0.8) & 75.4\% ($\pm$ 1.0) \\
\mmtsupf & 89.8\% ($\pm$ 0.6) & 89.4\% ($\pm$ 0.7) &  89.0\% ($\pm$ 2.0) \\
\mmtsup & 92.2\% ($\pm$ 0.2) & 92.3\% ($\pm$ 0.1) & 91.9\% ($\pm$ 0.2) \\
\hline
\end{tabular}}
\caption{Grounding ability of \textsc{wait-1} simultaneous MMT systems using features from the ground-truth region annotations. The scores presented are intersection over unions (IoU) on the \T{test2016} set, averaged across three runs.}
\label{tbl:mmt_grounding}
\end{table}





\begin{table}[t]
\centering
\def\arraystretch{1.2}
\resizebox{.99\textwidth}{!}{%
\begin{tabular}{l|lcr|lcr|lcr}
\hline
 & \MC{3}{c|}{\textsc{En-Cs}} & \MC{3}{c|}{\textsc{En-De}} & \MC{3}{c}{\textsc{En-Fr}} \\
 & \textsc{IoU} & \textsc{Cos} & \textsc{Em} & 
 \textsc{IoU} & \textsc{Cos} & \textsc{Em} &
 \textsc{IoU} & \textsc{Cos} & \textsc{Em} \\
\hline
\textsc{Mmt} & 23.1\% & 0.416 & 9.8\%    & 20.6\%& 0.403 & 8.4\%     & 28.5\%& 0.488 & 18.3\%  \\
\mmtsupf     & 42.5\% & 0.581 & 26.6\%   & 42.0\%& 0.578 & 26.1\%    & 41.4\% & 0.572 & 25.4\% \\
\mmtsup      & 44.3\% & 0.593 & 27.6\%   & 44.3\%& 0.590 & 27.4\%    & 44.4\%& 0.589 & 27.2\%  \\
\hline
\end{tabular}}
\caption{Grounding ability of \textsc{wait-1} simultaneous MMT systems using the 36 BUTD region proposals: \textsc{IoU, Cos} and \textsc{Em} denote intersection-over-union, cosine similarity and exact match accuracy, respectively. The scores are obtained from the \T{test2016} test set and averaged across three runs.}
\label{tbl:mmt_grounding_butd_cos}
\end{table}

Table~\ref{tbl:mmt_grounding_butd_cos} reports all three metrics averaged across three runs of each model. We observe that all three metrics show the same trend across model types. Regardless of the variant type, the incorporation of attention supervision dramatically improves the ratio of exact matches by up to $17\%$ in absolute terms. We also see that, unlike the previous experiment where the models were provided the human-labelled regions (Table~\ref{tbl:mmt_grounding}), here we have at least the \B{English\ra French} MMT model with unsupervised visual attention which obtains much better grounding scores than the other two language directions. Now that the models are given way more
regions (36) than the human-labelled scenario ($\sim$4.3), we think these results are more representative for the final grounding ability of these models. As expected, these results are a lot lower than Table~\ref{tbl:mmt_grounding} since (i) we had a lot fewer bounding boxes per image in that case and (ii), the BUTD bounding boxes are different from the gold ones, meaning that even if the model focused on a similar part of the image, the IoU score will be lower than 1.

\subsubsection{Visualising the Attention}
\begin{figure}[tp]
\centering
\includegraphics[width=0.99\textwidth]{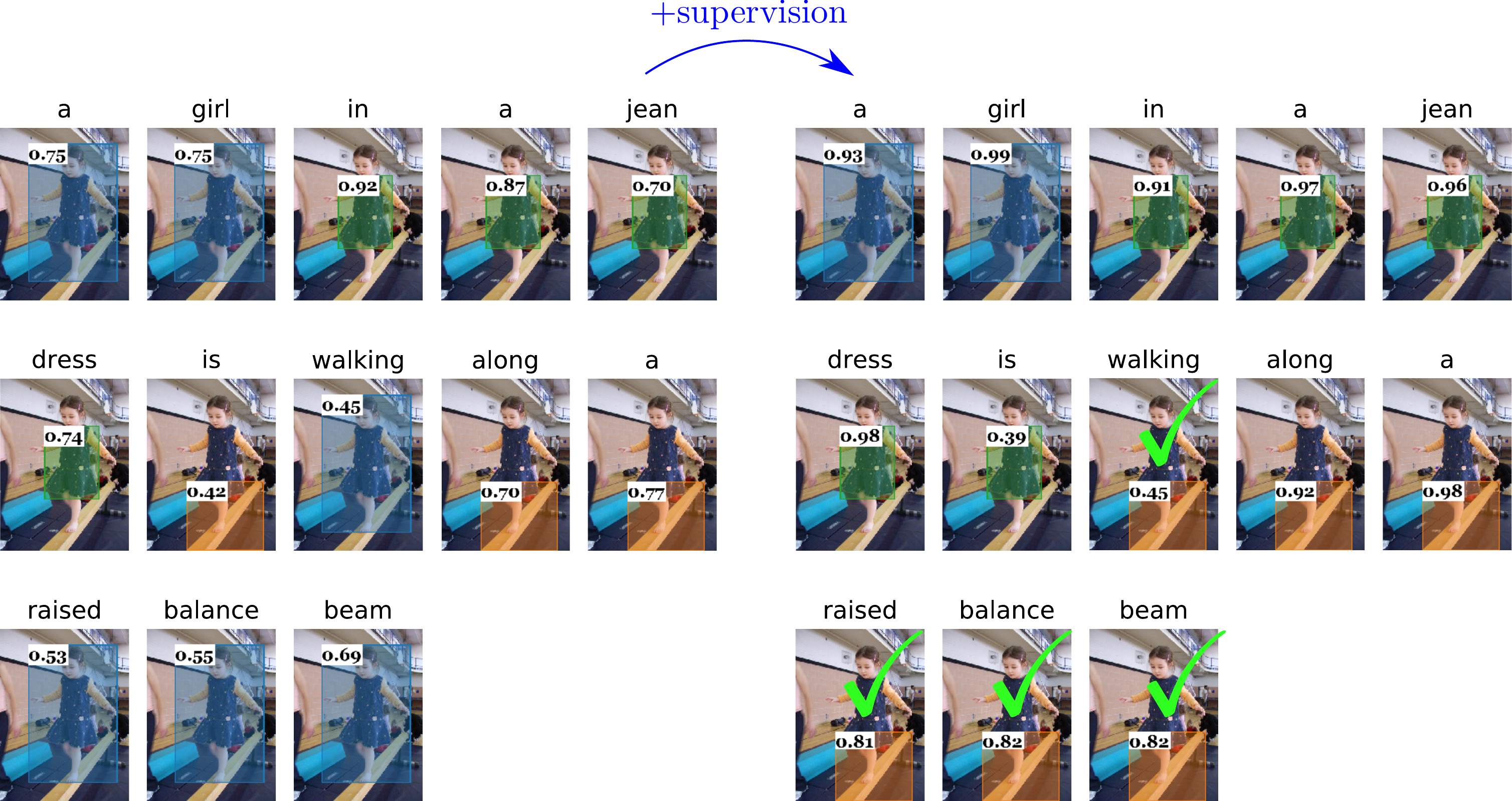}\\[1.7em]
\textcolor{gray}{\rule[1ex]{0.99\textwidth}{.3pt}}\\[1.7em]
\includegraphics[width=0.99\textwidth]{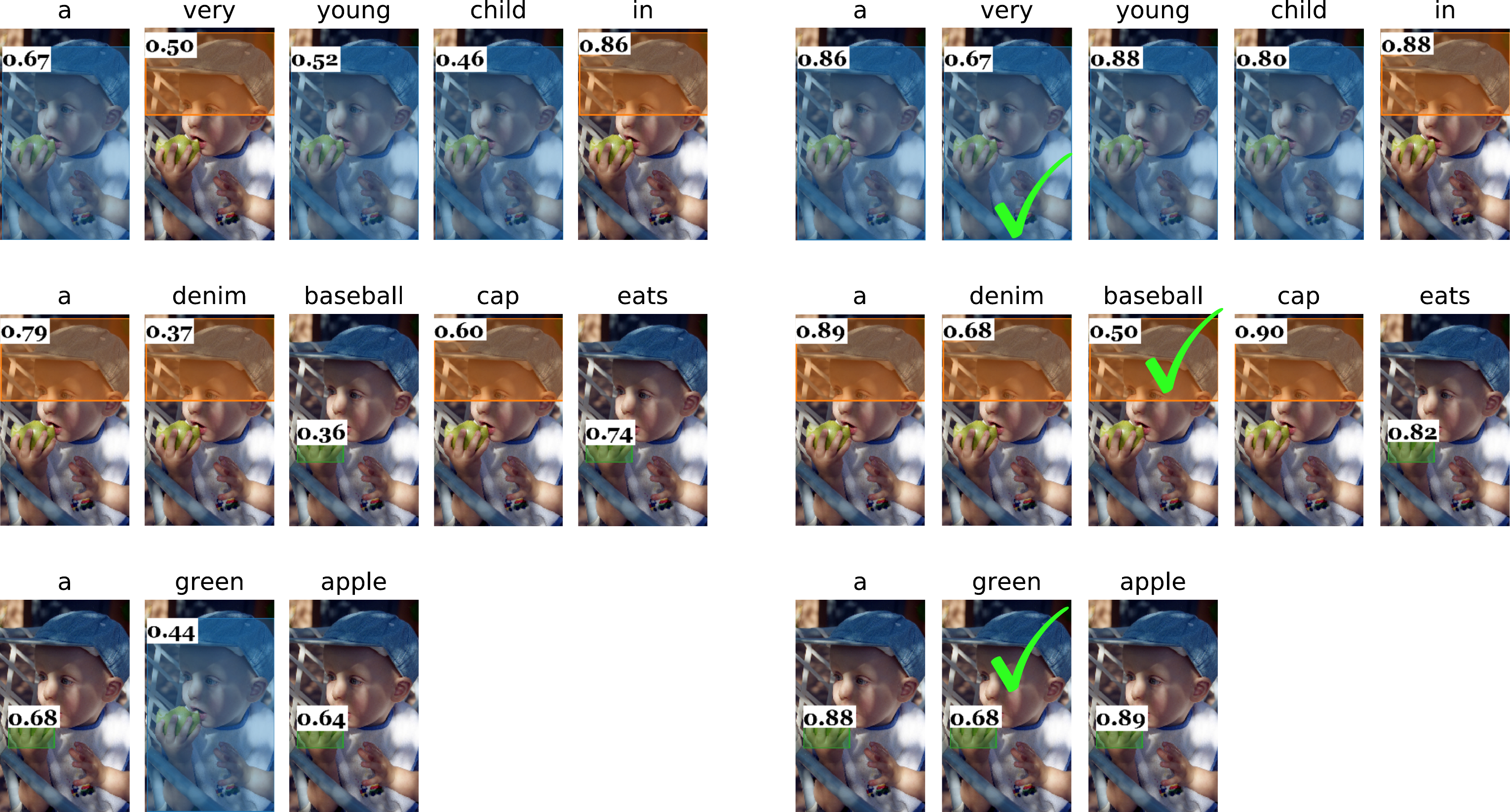}
\caption{Regional attention examples produced by the MMT model with unsupervised visual attention (left) and the \textsc{Fine-Tuned} variant (right): Green check marks show cases where supervision fixes incorrect word \& region alignments.}
\label{fig:attn_examples}
\end{figure}





Our main motivation to supervise the visual attention was to guide the model on learning cross-modal visual-linguistic alignments. After quantifying their ability to do so using metrics such as IoU and cosine similarity, we now visualise the impact of supervision on two real examples. Figure~\ref{fig:attn_examples} shows that the MMT models with unsupervised visual attention (on the left)
are already good at attending on relevant regions to some extent. However, there are still cases of misalignment: in the first example, none of the words in the phrase ``\I{raised balance beam}'' are aligned with the actual region showing the \I{beam}. Similar skews also happen in the second example such as the word \I{baseball} getting aligned with the region \I{apple}. The \textsc{Fine-Tuned} variants (on the right) not only fix these cases of misalignments but also provide much more confident attention distributions. This is particularly evident in the attention probabilities of nouns which, unlike connectives or verbs, have corresponding ground-truth bounding boxes.

\begin{figure}[b!]
\centering
\resizebox{.99\textwidth}{!}{%
\begin{tabular}{crl}
\toprule
\MR{9}{*}{\includegraphics[height=3cm]{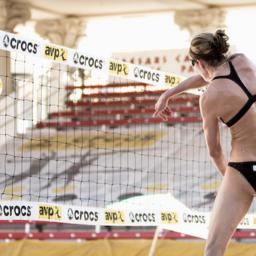}}
& \textsc{Src} & a woman is playing volleyball.  \\
& \textsc{Nmt} & \W{muž se dívá} na \C{volejbal.} \\
&              & \G{(man looks at volleyball.)} \\
& \textsc{Mmt} & \C{žena} v bílém oblečení \C{hraje volejbal.}\\
&              & \G{(woman in white clothes playing volleyball.)} \\
& \textsc{Sup} & \C{žena} \W{jde po ulici} \C{hraje volejbal.}\\
&              & \G{(woman goes down the street playing volleyball.)} \\
& \textsc{Sup+}& \C{žena, která hraje volejbal.} \\
&              & \G{(woman playing volleyball.)} \\
\cmidrule{1-3}
\MR{9}{*}{\includegraphics[height=3cm]{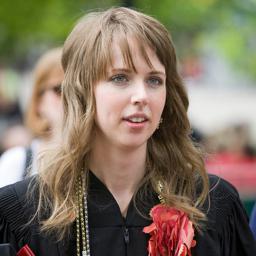}}
& \textsc{Src} & a woman with long hair is at a graduation ceremony.  \\
& \textsc{Nmt} & \W{ein} \C{frau mit langen haaren sitzt} bei einem \W{olympischen spielen bei einer konferenz.} \\
&              & \G{(a woman with long hair sitting at an olympic game at a conference.)} \\
& \textsc{Mmt} & \W{ein} \C{frau mit langen haaren ist an einem} \W{partner-konzert.} \\
&              & \G{(a woman with a long hair is in a partner concert.)} \\
& \textsc{Sup} & \C{eine frau mit langen haaren hält sich an einer} \W{konferenz} vor. \\
&              & \G{(a woman with a long hair is sitting in front of a conference.)} \\
& \textsc{Sup+}& \C{eine frau mit langen haaren steht an einem} \W{lutscher-bahnhof.} \\
&              & \G{(a woman with a long hair stands at a German railway.)} \\
\cmidrule{1-3}
\MR{9}{*}{\includegraphics[height=3cm]{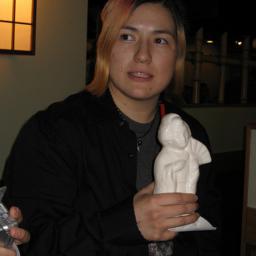}}
& \textsc{Src} & a woman is holding a small white statue.  \\
& \textsc{Nmt} & \W{un} \C{femme est en train de tenir une petite statue blanche.} \\
&              & \G{(a woman is holding a small white statue.)} \\
& \textsc{Mmt} & \W{un} \C{femme} \W{fait} de la \C{petite} \W{femme} \C{blanche.} \\
&              & \G{(a woman makes the little white woman.)} \\
& \textsc{Sup} & \C{une femme} fait un signe à un \C{petit} \W{homme} \C{blanc.} \\
&              & \G{(a woman makes a sign to a small white man.)} \\
& \textsc{Sup+}& \C{une femme est en train de tenir une petite statue blanche.}\\
&              & \G{(a woman is holding a small white statue.)} \\
\bottomrule
\end{tabular}}
\caption{\textsc{wait-1} translation examples from all model variants: \textsc{Sup} and \textsc{Sup+} denote the \textsc{Scratch} and \textsc{Fine-Tuned} variants of attention supervision, respectively. \C{blue} and \W{red} colors indicate correct and wrong lexical choices. ``Google Translate'' output for each non-English hypothesis is shown in grey.}
\label{fig:mmt_ex_wait1}
\end{figure}
\subsection{Translation Examples}
Finally, we now present qualitative translation examples from our simultaneous MT systems, to highlight both the strengths and weaknesses of the models explored.
Figure~\ref{fig:mmt_ex_wait1} shows an example for each language direction, using the \textsc{wait-1} decoded NMT, MMT and supervised MMT systems. For the first example in \B{Czech}, we see that the NMT system hopelessly relies on dataset biases to always generate ``Muž (man)'' when the definite article ``a'' is the only observed word at source side. Although the MMT with unsupervised visual attention and the supervision from scratch (\textsc{Sup}) both pick the correct translation ``Žena (woman)'', they somewhat hallucinate the continuation. In contrast, the fine-tuned  supervision (\textsc{Sup+}) generates the correct translation without any errors.
For the \B{German} example, all models struggle to translate ``graduation ceremony'' into German. Interestingly, they all pick words that refer to places such as ``concert'' or ``conference'' rather than purely random lexical choices, hinting at the fact that the models may be relying more on visual information than the language signal. In addition, NMT and the non-supervised MMT also commit an error when translating ``a'' into German \ie they both pick the masculine form ``ein'' instead of ``eine''. The supervised models are better to integrate visual features in this case, as they correctly pick ``eine''.
\begin{figure}[b!]
\centering
\resizebox{.99\textwidth}{!}{%
\begin{tabular}{cll}
\toprule
& &  \\
\MR{5}{*}{\includegraphics[height=3cm]{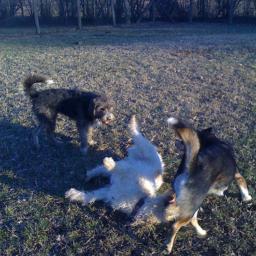}}
& \textsc{Src} & three dogs play with each other out in the field. \\
& \textsc{Sup} & \C{tři psi si spolu hrají} \W{ve sněhu se třemi psy.} \\
&              & \G{(three dogs playing together in the snow with three dogs.)} \\
& \textsc{Sup+}& \C{tři psi si spolu hrají venku na poli.} \\
&              & \G{(three dogs playing together outside in the field.)} \\
& &  \\
& &  \\
\cmidrule{1-3}
& &  \\
\MR{5}{*}{\includegraphics[height=3cm]{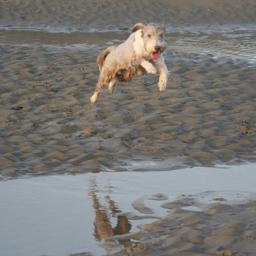}}
& \textsc{Src} & a gray and white dog jumping over standing water in the sand. \\
& \textsc{Sup} & \C{ein grau-weißer hund} \W{steht im wasser und springt über das wasser.}\\
&              & \G{(a gray and white dog stands in the water and jumps over the water.)} \\
& \textsc{Sup+}& \C{ein grau-weißer hund springt über das wasser im sand.} \\
&              & \G{(a gray and white dog leaps over the water in the sand.)} \\
& &  \\
& &  \\
\cmidrule{1-3}
& &  \\
\MR{5}{*}{\includegraphics[height=3cm]{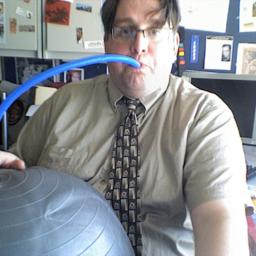}}
& \textsc{Src} & a man is blowing into a plastic ball. \\
& \textsc{Sup} & \C{un homme} \W{porte des lunettes} \C{dans un ballon en plastique.} \\
&              & \G{(a man wears glasses in a plastic ball.)} \\
& \textsc{Sup+}& \C{un homme est en train de souffler dans une balle en plastique.}\\
&              & \G{(a man is blowing into a plastic ball.)} \\
& &  \\
& &  \\
\bottomrule
\end{tabular}}
\caption{\textsc{wait-3} translation examples comparing two variants of attention supervision: \textsc{Sup} and \textsc{Sup+} denote the \textsc{Scratch} and \textsc{Fine-Tuned} variants, respectively. \C{blue} and \W{red} colors indicate correct and wrong lexical choices. ``Google Translate'' output for each non-English hypothesis is shown in grey.}
\label{fig:mmt_ex_wait3}
\end{figure}
Finally, the \B{French} example shows similar patterns to the German one as well: only the supervised models are able to pick the correct article form ``une'' and it is the fine-tuned variant that translates the whole sentence in a correct way. If we further look at examples of \textsc{wait-3} simultaneous MT models and contrast the \textsc{Scratch} and \textsc{Fine-Tuned} variants (Figure~\ref{fig:mmt_ex_wait3}), we observe that both variants are more or less equivalent when translating the early parts of the sentences but the \textsc{Scratch} models commit critical errors in overall, when the final translations are taken into account.



\section{Conclusions}
\label{sec:conclusion}
In this paper, we proposed the first Transformer-based simultaneous multimodal MT model and extended it with an auxiliary supervision signal that guides the visual attention mechanism using ground-truth phrase-region alignments. We performed extensive experiments on the three language pairs of the Multi30k dataset and upon conducting thorough quantitative and qualitative analyses, we showed that (i) supervised visual attention consistently improves the translation quality of the MMT models, and (ii) fine-tuning the MMT with the supervision loss results in better performance than training the MMT from scratch.

\section*{Acknowledgments}
This article follows from the MSc. Thesis by Veneta Haralampieva, co-supervised by Lucia Specia and Ozan Caglayan. Lucia Specia and Ozan Caglayan were supported by the MultiMT project (H2020 ERC Starting Grant No. 678017). Lucia Specia also received support from the Air Force Office of Scientific Research (under award number FA8655-20-1-7006).

\bibliography{paper}

\begin{thebibliography}{}

\bibitem[\protect\BCAY{Alinejad, Siahbani,\ \BBA\ Sarkar}{Alinejad
  et~al.}{2018}]{alinejad2018prediction}
Alinejad, A., Siahbani, M., \BBA\ Sarkar, A. \BBOP2018\BBCP.
\newblock \BBOQ Prediction improves simultaneous neural machine
  translation\BBCQ\
\newblock In {\Bem Proceedings of the 2018 Conference on Empirical Methods in
  Natural Language Processing}, \BPGS\ 3022--3027, Brussels, Belgium.
  Association for Computational Linguistics.

\bibitem[\protect\BCAY{Anderson, He, Buehler, Teney, Johnson, Gould,\ \BBA\
  Zhang}{Anderson et~al.}{2018}]{Anderson2017up-down}
Anderson, P., He, X., Buehler, C., Teney, D., Johnson, M., Gould, S., \BBA\
  Zhang, L. \BBOP2018\BBCP.
\newblock \BBOQ Bottom-up and top-down attention for image captioning and
  visual question answering\BBCQ\
\newblock In {\Bem CVPR}.

\bibitem[\protect\BCAY{Arivazhagan, Cherry, Macherey, Chiu, Yavuz, Pang, Li,\
  \BBA\ Raffel}{Arivazhagan et~al.}{2019}]{arivazhagan2019monotonic}
Arivazhagan, N., Cherry, C., Macherey, W., Chiu, C.-C., Yavuz, S., Pang, R.,
  Li, W., \BBA\ Raffel, C. \BBOP2019\BBCP.
\newblock \BBOQ Monotonic infinite lookback attention for simultaneous machine
  translation\BBCQ\
\newblock In {\Bem Proceedings of the 57th Annual Meeting of the Association
  for Computational Linguistics}, \BPGS\ 1313--1323.

\bibitem[\protect\BCAY{Arivazhagan, Cherry, Macherey,\ \BBA\
  Foster}{Arivazhagan et~al.}{2020}]{arivazhagan2020re}
Arivazhagan, N., Cherry, C., Macherey, W., \BBA\ Foster, G. \BBOP2020\BBCP.
\newblock \BBOQ Re-translation versus streaming for simultaneous
  translation\BBCQ\
\newblock In {\Bem Proceedings of the 17th International Conference on Spoken
  Language Translation}, \BPGS\ 220--227, Online. Association for Computational
  Linguistics.

\bibitem[\protect\BCAY{Arthur, Cohn,\ \BBA\ Haffari}{Arthur
  et~al.}{2021}]{arthur2021learning}
Arthur, P., Cohn, T., \BBA\ Haffari, G. \BBOP2021\BBCP.
\newblock \BBOQ Learning coupled policies for simultaneous machine translation
  using imitation learning\BBCQ\
\newblock In {\Bem Proceedings of the 16th Conference of the European Chapter
  of the Association for Computational Linguistics: Main Volume}, \BPGS\
  2709--2719, Online. Association for Computational Linguistics.

\bibitem[\protect\BCAY{Ba, Kiros,\ \BBA\ Hinton}{Ba
  et~al.}{2016}]{ba2016layernorm}
Ba, J.~L., Kiros, J.~R., \BBA\ Hinton, G.~E. \BBOP2016\BBCP.
\newblock \BBOQ Layer normalization\BBCQ\
\newblock {\Bem arXiv preprint arXiv:1607.06450}, {\Bem 1\/}(1).

\bibitem[\protect\BCAY{Bahdanau, Cho,\ \BBA\ Bengio}{Bahdanau
  et~al.}{2015}]{Bahdanau2015NeuralMT}
Bahdanau, D., Cho, K., \BBA\ Bengio, Y. \BBOP2015\BBCP.
\newblock \BBOQ Neural machine translation by jointly learning to align and
  translate\BBCQ\
\newblock In {\Bem Proceedings of the 3rd International Conference on Learning
  Representations}.

\bibitem[\protect\BCAY{Bangalore, Rangarajan~Sridhar, Kolan, Golipour,\ \BBA\
  Jimenez}{Bangalore et~al.}{2012}]{bangalore-etal-2012-real}
Bangalore, S., Rangarajan~Sridhar, V.~K., Kolan, P., Golipour, L., \BBA\
  Jimenez, A. \BBOP2012\BBCP.
\newblock \BBOQ Real-time incremental speech-to-speech translation of
  dialogs\BBCQ\
\newblock In {\Bem Proceedings of the 2012 Conference of the North {A}merican
  Chapter of the Association for Computational Linguistics: Human Language
  Technologies}, \BPGS\ 437--445, Montr{\'e}al, Canada. Association for
  Computational Linguistics.

\bibitem[\protect\BCAY{Barrault, Bougares, Specia, Lala, Elliott,\ \BBA\
  Frank}{Barrault et~al.}{2018}]{barrault-etal-2018-findings}
Barrault, L., Bougares, F., Specia, L., Lala, C., Elliott, D., \BBA\ Frank, S.
  \BBOP2018\BBCP.
\newblock \BBOQ Findings of the third shared task on multimodal machine
  translation\BBCQ\
\newblock In {\Bem Proceedings of the Third Conference on Machine Translation,
  Volume 2: Shared Task Papers}, \BPGS\ 308--327, Belgium, Brussels.
  Association for Computational Linguistics.

\bibitem[\protect\BCAY{Bird\ \BBA\ Loper}{Bird\ \BBA\
  Loper}{2004}]{bird-loper-2004-nltk}
Bird, S.\BBACOMMA\  \BBA\ Loper, E. \BBOP2004\BBCP.
\newblock \BBOQ {NLTK}: The natural language toolkit\BBCQ\
\newblock In {\Bem Proceedings of the {ACL} Interactive Poster and
  Demonstration Sessions}, \BPGS\ 214--217, Barcelona, Spain. Association for
  Computational Linguistics.

\bibitem[\protect\BCAY{Bub, Wahlster,\ \BBA\ Waibel}{Bub
  et~al.}{1997}]{bub1997verbmobil}
Bub, T., Wahlster, W., \BBA\ Waibel, A. \BBOP1997\BBCP.
\newblock \BBOQ Verbmobil: The combination of deep and shallow processing for
  spontaneous speech translation\BBCQ\
\newblock In {\Bem 1997 IEEE International Conference on Acoustics, Speech, and
  Signal Processing}, \lowercase{\BVOL}~1, \BPGS\ 71--74. IEEE.

\bibitem[\protect\BCAY{Caglayan}{Caglayan}{2019}]{caglayan-thesis-2019}
Caglayan, O. \BBOP2019\BBCP.
\newblock {\Bem {Multimodal Machine Translation}}.
\newblock Theses, {Universit{\'e} du Maine}.

\bibitem[\protect\BCAY{Caglayan, Aransa, Bardet, Garc{\'\i}a-Mart{\'\i}nez,
  Bougares, Barrault, Masana, Herranz,\ \BBA\ van~de Weijer}{Caglayan
  et~al.}{2017}]{caglayan2017lium}
Caglayan, O., Aransa, W., Bardet, A., Garc{\'\i}a-Mart{\'\i}nez, M., Bougares,
  F., Barrault, L., Masana, M., Herranz, L., \BBA\ van~de Weijer, J.
  \BBOP2017\BBCP.
\newblock \BBOQ {LIUM-CVC Submissions for WMT17 Multimodal Translation
  Task}\BBCQ\
\newblock In {\Bem Proceedings of the Second Conference on Machine
  Translation}, \BPGS\ 432--439.

\bibitem[\protect\BCAY{Caglayan, Aransa, Wang, Masana,
  Garc{\'\i}a-Mart{\'\i}nez, Bougares, Barrault,\ \BBA\ van~de Weijer}{Caglayan
  et~al.}{2016}]{caglayan2016does}
Caglayan, O., Aransa, W., Wang, Y., Masana, M., Garc{\'\i}a-Mart{\'\i}nez, M.,
  Bougares, F., Barrault, L., \BBA\ van~de Weijer, J. \BBOP2016\BBCP.
\newblock \BBOQ Does multimodality help human and machine for translation and
  image captioning?\BBCQ\
\newblock In {\Bem Proceedings of the First Conference on Machine Translation:
  Volume 2, Shared Task Papers}, \BPGS\ 627--633, Berlin, Germany. Association
  for Computational Linguistics.

\bibitem[\protect\BCAY{Caglayan, Ive, Haralampieva, Madhyastha, Barrault,\
  \BBA\ Specia}{Caglayan et~al.}{2020a}]{caglayan2020simultaneous}
Caglayan, O., Ive, J., Haralampieva, V., Madhyastha, P., Barrault, L., \BBA\
  Specia, L. \BBOP2020a\BBCP.
\newblock \BBOQ Simultaneous machine translation with visual context\BBCQ\
\newblock In {\Bem Proceedings of the 2020 Conference on Empirical Methods in
  Natural Language Processing (EMNLP)}, \BPGS\ 2350--2361, Online. Association
  for Computational Linguistics.

\bibitem[\protect\BCAY{Caglayan, Madhyastha,\ \BBA\ Specia}{Caglayan
  et~al.}{2020b}]{caglayan-etal-2020-curious}
Caglayan, O., Madhyastha, P., \BBA\ Specia, L. \BBOP2020b\BBCP.
\newblock \BBOQ Curious case of language generation evaluation metrics: A
  cautionary tale\BBCQ\
\newblock In {\Bem Proceedings of the 28th International Conference on
  Computational Linguistics}, \BPGS\ 2322--2328, Barcelona, Spain (Online).
  International Committee on Computational Linguistics.

\bibitem[\protect\BCAY{Calixto, Elliott,\ \BBA\ Frank}{Calixto
  et~al.}{2016}]{calixto2016dcu}
Calixto, I., Elliott, D., \BBA\ Frank, S. \BBOP2016\BBCP.
\newblock \BBOQ {DCU-UvA multimodal MT system report}\BBCQ\
\newblock In {\Bem Proceedings of the First Conference on Machine Translation:
  Volume 2, Shared Task Papers}, \BPGS\ 634--638.

\bibitem[\protect\BCAY{Calixto\ \BBA\ Liu}{Calixto\ \BBA\
  Liu}{2017}]{calixto2017incorporating}
Calixto, I.\BBACOMMA\  \BBA\ Liu, Q. \BBOP2017\BBCP.
\newblock \BBOQ Incorporating global visual features into attention-based
  neural machine translation.\BBCQ\
\newblock In {\Bem Proceedings of the 2017 Conference on Empirical Methods in
  Natural Language Processing}, \BPGS\ 992--1003.

\bibitem[\protect\BCAY{Calixto, Liu,\ \BBA\ Campbell}{Calixto
  et~al.}{2017}]{calixto2017doubly}
Calixto, I., Liu, Q., \BBA\ Campbell, N. \BBOP2017\BBCP.
\newblock \BBOQ Doubly-attentive decoder for multi-modal neural machine
  translation\BBCQ\
\newblock In {\Bem Proceedings of the 55th Annual Meeting of the Association
  for Computational Linguistics (Volume 1: Long Papers)}, \BPGS\ 1913--1924.

\bibitem[\protect\BCAY{Chen, Fang, Lin, Vedantam, Gupta, Doll{\'a}r,\ \BBA\
  Zitnick}{Chen et~al.}{2015}]{chen-etal-2015-coco}
Chen, X., Fang, H., Lin, T.-Y., Vedantam, R., Gupta, S., Doll{\'a}r, P., \BBA\
  Zitnick, C.~L. \BBOP2015\BBCP.
\newblock \BBOQ {Microsoft COCO captions: Data collection and evaluation
  server}\BBCQ\
\newblock {\Bem arXiv preprint arXiv:1504.00325}, {\Bem 1}.

\bibitem[\protect\BCAY{Cho\ \BBA\ Esipova}{Cho\ \BBA\
  Esipova}{2016}]{cho2016can}
Cho, K.\BBACOMMA\  \BBA\ Esipova, M. \BBOP2016\BBCP.
\newblock \BBOQ Can neural machine translation do simultaneous
  translation?\BBCQ\
\newblock {\Bem arXiv preprint arXiv:1606.02012}, {\Bem 1}.

\bibitem[\protect\BCAY{Dalvi, Durrani, Sajjad,\ \BBA\ Vogel}{Dalvi
  et~al.}{2018}]{dalvi2018incremental}
Dalvi, F., Durrani, N., Sajjad, H., \BBA\ Vogel, S. \BBOP2018\BBCP.
\newblock \BBOQ Incremental decoding and training methods for simultaneous
  translation in neural machine translation\BBCQ\
\newblock In {\Bem Proceedings of the 2018 Conference of the North {A}merican
  Chapter of the Association for Computational Linguistics: Human Language
  Technologies, Volume 2 (Short Papers)}, \BPGS\ 493--499, New Orleans,
  Louisiana. Association for Computational Linguistics.

\bibitem[\protect\BCAY{Delbrouck\ \BBA\ Dupont}{Delbrouck\ \BBA\
  Dupont}{2017}]{delbrouck2017modulating}
Delbrouck, J.-B.\BBACOMMA\  \BBA\ Dupont, S. \BBOP2017\BBCP.
\newblock \BBOQ Modulating and attending the source image during encoding
  improves multimodal translation\BBCQ\
\newblock {\Bem arXiv preprint arXiv:1712.03449}, {\Bem 1\/}(1).

\bibitem[\protect\BCAY{Deng, Dong, Socher, Li, Li,\ \BBA\ Fei-Fei}{Deng
  et~al.}{2009}]{deng2009imagenet}
Deng, J., Dong, W., Socher, R., Li, L.-J., Li, K., \BBA\ Fei-Fei, L.
  \BBOP2009\BBCP.
\newblock \BBOQ Imagenet: A large-scale hierarchical image database\BBCQ\
\newblock In {\Bem 2009 IEEE conference on computer vision and pattern
  recognition}, \BPGS\ 248--255. Ieee.

\bibitem[\protect\BCAY{Denkowski\ \BBA\ Lavie}{Denkowski\ \BBA\
  Lavie}{2014}]{meteor}
Denkowski, M.\BBACOMMA\  \BBA\ Lavie, A. \BBOP2014\BBCP.
\newblock \BBOQ Meteor universal: Language specific translation evaluation for
  any target language\BBCQ\
\newblock In {\Bem Proceedings of the Ninth Workshop on Statistical Machine
  Translation}, \BPGS\ 376--380. Association for Computational Linguistics.

\bibitem[\protect\BCAY{Elbayad, Besacier,\ \BBA\ Verbeek}{Elbayad
  et~al.}{2020}]{elbayad2020efficient}
Elbayad, M., Besacier, L., \BBA\ Verbeek, J. \BBOP2020\BBCP.
\newblock \BBOQ {Efficient Wait-k Models for Simultaneous Machine
  Translation}\BBCQ\
\newblock In {\Bem Proc. Interspeech 2020}, \BPGS\ 1461--1465.

\bibitem[\protect\BCAY{Elliott, Frank, Barrault, Bougares,\ \BBA\
  Specia}{Elliott et~al.}{2017}]{elliott-etal-2017-findings}
Elliott, D., Frank, S., Barrault, L., Bougares, F., \BBA\ Specia, L.
  \BBOP2017\BBCP.
\newblock \BBOQ Findings of the second shared task on multimodal machine
  translation and multilingual image description\BBCQ\
\newblock In {\Bem Proceedings of the Second Conference on Machine Translation,
  Volume 2: Shared Task Papers}, \BPGS\ 215--233, Copenhagen, Denmark.
  Association for Computational Linguistics.

\bibitem[\protect\BCAY{Elliott, Frank, Sima{'}an,\ \BBA\ Specia}{Elliott
  et~al.}{2016}]{elliott-etal-2016-multi30k}
Elliott, D., Frank, S., Sima{'}an, K., \BBA\ Specia, L. \BBOP2016\BBCP.
\newblock \BBOQ {M}ulti30{K}: Multilingual {E}nglish-{G}erman image
  descriptions\BBCQ\
\newblock In {\Bem Proceedings of the 5th Workshop on Vision and Language},
  \BPGS\ 70--74, Berlin, Germany. Association for Computational Linguistics.

\bibitem[\protect\BCAY{Elliott\ \BBA\ K{\'a}d{\'a}r}{Elliott\ \BBA\
  K{\'a}d{\'a}r}{2017}]{elliott-kadar-2017-imagination}
Elliott, D.\BBACOMMA\  \BBA\ K{\'a}d{\'a}r, {\'A}. \BBOP2017\BBCP.
\newblock \BBOQ Imagination improves multimodal translation\BBCQ\
\newblock In {\Bem Proceedings of the Eighth International Joint Conference on
  Natural Language Processing (Volume 1: Long Papers)}, \BPGS\ 130--141,
  Taipei, Taiwan. Asian Federation of Natural Language Processing.

\bibitem[\protect\BCAY{Garg, Peitz, Nallasamy,\ \BBA\ Paulik}{Garg
  et~al.}{2019}]{garg2019jointly}
Garg, S., Peitz, S., Nallasamy, U., \BBA\ Paulik, M. \BBOP2019\BBCP.
\newblock \BBOQ Jointly learning to align and translate with transformer
  models\BBCQ\
\newblock In {\Bem Conference on Empirical Methods in Natural Language
  Processing (EMNLP)}, Hong Kong.

\bibitem[\protect\BCAY{Gu, Neubig, Cho,\ \BBA\ Li}{Gu
  et~al.}{2017}]{gu2016learning}
Gu, J., Neubig, G., Cho, K., \BBA\ Li, V.~O. \BBOP2017\BBCP.
\newblock \BBOQ Learning to translate in real-time with neural machine
  translation\BBCQ\
\newblock In {\Bem Proceedings of the 15th Conference of the {E}uropean Chapter
  of the Association for Computational Linguistics: Volume 1, Long Papers},
  \BPGS\ 1053--1062, Valencia, Spain. Association for Computational
  Linguistics.

\bibitem[\protect\BCAY{He, Zhang, Ren,\ \BBA\ Sun}{He
  et~al.}{2016}]{he2016deep}
He, K., Zhang, X., Ren, S., \BBA\ Sun, J. \BBOP2016\BBCP.
\newblock \BBOQ Deep residual learning for image recognition\BBCQ\
\newblock In {\Bem Proceedings of the IEEE conference on computer vision and
  pattern recognition}, \BPGS\ 770--778.

\bibitem[\protect\BCAY{Imankulova, Kaneko, Hirasawa,\ \BBA\ Komachi}{Imankulova
  et~al.}{2020}]{imankulova-etal-2020-towards}
Imankulova, A., Kaneko, M., Hirasawa, T., \BBA\ Komachi, M. \BBOP2020\BBCP.
\newblock \BBOQ Towards multimodal simultaneous neural machine
  translation\BBCQ\
\newblock In {\Bem Proceedings of the Fifth Conference on Machine Translation},
  \BPGS\ 594--603, Online. Association for Computational Linguistics.

\bibitem[\protect\BCAY{Ive, Li, Miao, Caglayan, Madhyastha,\ \BBA\ Specia}{Ive
  et~al.}{2021}]{ive-etal-2021-exploiting}
Ive, J., Li, A.~M., Miao, Y., Caglayan, O., Madhyastha, P., \BBA\ Specia, L.
  \BBOP2021\BBCP.
\newblock \BBOQ Exploiting multimodal reinforcement learning for simultaneous
  machine translation\BBCQ\
\newblock In {\Bem Proceedings of the 16th Conference of the European Chapter
  of the Association for Computational Linguistics: Main Volume}, \BPGS\
  3222--3233, Online. Association for Computational Linguistics.

\bibitem[\protect\BCAY{Kingma\ \BBA\ Ba}{Kingma\ \BBA\
  Ba}{2014}]{kingma2014adam}
Kingma, D.~P.\BBACOMMA\  \BBA\ Ba, J. \BBOP2014\BBCP.
\newblock \BBOQ Adam: A method for stochastic optimization\BBCQ\
\newblock {\Bem arXiv preprint arXiv:1412.6980}, {\Bem 1}.

\bibitem[\protect\BCAY{Koehn, Hoang, Birch, Callison-Burch, Federico, Bertoldi,
  Cowan, Shen, Moran, Zens, Dyer, Bojar, Constantin,\ \BBA\ Herbst}{Koehn
  et~al.}{2007}]{koehn-etal-2007-moses}
Koehn, P., Hoang, H., Birch, A., Callison-Burch, C., Federico, M., Bertoldi,
  N., Cowan, B., Shen, W., Moran, C., Zens, R., Dyer, C., Bojar, O.,
  Constantin, A., \BBA\ Herbst, E. \BBOP2007\BBCP.
\newblock \BBOQ {M}oses: Open source toolkit for statistical machine
  translation\BBCQ\
\newblock In {\Bem Proceedings of the 45th Annual Meeting of the Association
  for Computational Linguistics Companion Volume Proceedings of the Demo and
  Poster Sessions}, \BPGS\ 177--180, Prague, Czech Republic. Association for
  Computational Linguistics.

\bibitem[\protect\BCAY{Krishna, Zhu, Groth, Johnson, Hata, Kravitz, Chen,
  Kalantidis, Li, Shamma, et~al.}{Krishna et~al.}{2017}]{krishna2017visual}
Krishna, R., Zhu, Y., Groth, O., Johnson, J., Hata, K., Kravitz, J., Chen, S.,
  Kalantidis, Y., Li, L.-J., Shamma, D.~A., et~al. \BBOP2017\BBCP.
\newblock \BBOQ Visual genome: Connecting language and vision using
  crowdsourced dense image annotations\BBCQ\
\newblock {\Bem International journal of computer vision}, {\Bem 123\/}(1),
  32--73.

\bibitem[\protect\BCAY{Libovick{\`y}\ \BBA\ Helcl}{Libovick{\`y}\ \BBA\
  Helcl}{2017}]{libovicky2017attention}
Libovick{\`y}, J.\BBACOMMA\  \BBA\ Helcl, J. \BBOP2017\BBCP.
\newblock \BBOQ Attention strategies for multi-source sequence-to-sequence
  learning\BBCQ\
\newblock In {\Bem Proceedings of the 55th Annual Meeting of the Association
  for Computational Linguistics (Volume 2: Short Papers)}, \BPGS\ 196--202.

\bibitem[\protect\BCAY{Libovick{\`y}, Helcl,\ \BBA\
  Mare{\v{c}}ek}{Libovick{\`y} et~al.}{2018}]{libovicky2018input}
Libovick{\`y}, J., Helcl, J., \BBA\ Mare{\v{c}}ek, D. \BBOP2018\BBCP.
\newblock \BBOQ Input combination strategies for multi-source transformer
  decoder\BBCQ\
\newblock In {\Bem Proceedings of the Third Conference on Machine Translation:
  Research Papers}, \BPGS\ 253--260.

\bibitem[\protect\BCAY{Liu, Utiyama, Finch,\ \BBA\ Sumita}{Liu
  et~al.}{2016}]{liu-etal-2016-neural}
Liu, L., Utiyama, M., Finch, A., \BBA\ Sumita, E. \BBOP2016\BBCP.
\newblock \BBOQ Neural machine translation with supervised attention\BBCQ\
\newblock In {\Bem Proceedings of {COLING} 2016, the 26th International
  Conference on Computational Linguistics: Technical Papers}, \BPGS\
  3093--3102, Osaka, Japan. The COLING 2016 Organizing Committee.

\bibitem[\protect\BCAY{Lu, Batra, Parikh,\ \BBA\ Lee}{Lu
  et~al.}{2019}]{lu2019vilbert}
Lu, J., Batra, D., Parikh, D., \BBA\ Lee, S. \BBOP2019\BBCP.
\newblock \BBOQ Vilbert: Pretraining task-agnostic visiolinguistic
  representations for vision-and-language tasks\BBCQ\
\newblock In {\Bem Advances in Neural Information Processing Systems}, \BPGS\
  13--23.

\bibitem[\protect\BCAY{Lu, Yang, Batra,\ \BBA\ Parikh}{Lu
  et~al.}{2016}]{lu2016hierarchical}
Lu, J., Yang, J., Batra, D., \BBA\ Parikh, D. \BBOP2016\BBCP.
\newblock \BBOQ Hierarchical question-image co-attention for visual question
  answering\BBCQ\
\newblock In {\Bem Advances in neural information processing systems}, \BPGS\
  289--297.

\bibitem[\protect\BCAY{Ma, Huang, Xiong, Zheng, Liu, Zheng, Zhang, He, Liu, Li,
  et~al.}{Ma et~al.}{2019}]{ma2019stacl}
Ma, M., Huang, L., Xiong, H., Zheng, R., Liu, K., Zheng, B., Zhang, C., He, Z.,
  Liu, H., Li, X., et~al. \BBOP2019\BBCP.
\newblock \BBOQ {STACL: Simultaneous Translation with Implicit Anticipation and
  Controllable Latency using Prefix-to-Prefix Framework}\BBCQ\
\newblock In {\Bem Proceedings of the 57th Annual Meeting of the Association
  for Computational Linguistics}, \BPGS\ 3025--3036.

\bibitem[\protect\BCAY{Ma, Pino, Cross, Puzon,\ \BBA\ Gu}{Ma
  et~al.}{2020}]{ma2019monotonic}
Ma, X., Pino, J.~M., Cross, J., Puzon, L., \BBA\ Gu, J. \BBOP2020\BBCP.
\newblock \BBOQ Monotonic multihead attention\BBCQ\
\newblock In {\Bem 8th International Conference on Learning Representations,
  {ICLR} 2020, Addis Ababa, Ethiopia, April 26-30, 2020}.

\bibitem[\protect\BCAY{Mi, Wang,\ \BBA\ Ittycheriah}{Mi
  et~al.}{2016}]{mi-etal-2016-supervised}
Mi, H., Wang, Z., \BBA\ Ittycheriah, A. \BBOP2016\BBCP.
\newblock \BBOQ Supervised attentions for neural machine translation\BBCQ\
\newblock In {\Bem Proceedings of the 2016 Conference on Empirical Methods in
  Natural Language Processing}, \BPGS\ 2283--2288, Austin, Texas. Association
  for Computational Linguistics.

\bibitem[\protect\BCAY{Niehues, Pham, Ha, Sperber,\ \BBA\ Waibel}{Niehues
  et~al.}{2018}]{niehues2018}
Niehues, J., Pham, N.-Q., Ha, T.-L., Sperber, M., \BBA\ Waibel, A.
  \BBOP2018\BBCP.
\newblock \BBOQ Low-latency neural speech translation\BBCQ\
\newblock In {\Bem Proc. Interspeech 2018}, \BPGS\ 1293--1297.

\bibitem[\protect\BCAY{Nishihara, Tamura, Ninomiya, Omote,\ \BBA\
  Nakayama}{Nishihara et~al.}{2020}]{nishihara-etal-2020-supervised}
Nishihara, T., Tamura, A., Ninomiya, T., Omote, Y., \BBA\ Nakayama, H.
  \BBOP2020\BBCP.
\newblock \BBOQ Supervised visual attention for multimodal neural machine
  translation\BBCQ\
\newblock In {\Bem Proceedings of the 28th International Conference on
  Computational Linguistics}, \BPGS\ 4304--4314, Barcelona, Spain (Online).
  International Committee on Computational Linguistics.

\bibitem[\protect\BCAY{Papineni, Roukos, Ward,\ \BBA\ Zhu}{Papineni
  et~al.}{2002}]{papineni2002bleu}
Papineni, K., Roukos, S., Ward, T., \BBA\ Zhu, W.-J. \BBOP2002\BBCP.
\newblock \BBOQ {BLEU: a method for automatic evaluation of machine
  translation}\BBCQ\
\newblock In {\Bem Proceedings of the 40th annual meeting on association for
  computational linguistics}, \BPGS\ 311--318. Association for Computational
  Linguistics.

\bibitem[\protect\BCAY{Pascanu, Gulcehre, Cho,\ \BBA\ Bengio}{Pascanu
  et~al.}{2014}]{pascanu2014construct}
Pascanu, R., Gulcehre, C., Cho, K., \BBA\ Bengio, Y. \BBOP2014\BBCP.
\newblock \BBOQ How to construct deep recurrent neural networks: Proceedings of
  the second international conference on learning representations (iclr
  2014)\BBCQ\
\newblock In {\Bem 2nd International Conference on Learning Representations,
  ICLR 2014}.

\bibitem[\protect\BCAY{Pennington, Socher,\ \BBA\ Manning}{Pennington
  et~al.}{2014}]{pennington-etal-2014-glove}
Pennington, J., Socher, R., \BBA\ Manning, C. \BBOP2014\BBCP.
\newblock \BBOQ {G}lo{V}e: Global vectors for word representation\BBCQ\
\newblock In {\Bem Proceedings of the 2014 Conference on Empirical Methods in
  Natural Language Processing ({EMNLP})}, \BPGS\ 1532--1543, Doha, Qatar.
  Association for Computational Linguistics.

\bibitem[\protect\BCAY{Plummer, Wang, Cervantes, Caicedo, Hockenmaier,\ \BBA\
  Lazebnik}{Plummer et~al.}{2015}]{plummer-etal-2015-collecting}
Plummer, B.~A., Wang, L., Cervantes, C.~M., Caicedo, J.~C., Hockenmaier, J.,
  \BBA\ Lazebnik, S. \BBOP2015\BBCP.
\newblock \BBOQ {Flickr30k Entities: Collecting Region-to-Phrase
  Correspondences for Richer Image-to-Sentence Models}\BBCQ\
\newblock In {\Bem 2015 IEEE International Conference on Computer Vision
  (ICCV)}, \BPGS\ 2641--2649.

\bibitem[\protect\BCAY{Press\ \BBA\ Wolf}{Press\ \BBA\
  Wolf}{2017}]{press2017using}
Press, O.\BBACOMMA\  \BBA\ Wolf, L. \BBOP2017\BBCP.
\newblock \BBOQ Using the output embedding to improve language models\BBCQ\
\newblock In {\Bem Proceedings of the 15th Conference of the European Chapter
  of the Association for Computational Linguistics: Volume 2, Short Papers},
  \BPGS\ 157--163.

\bibitem[\protect\BCAY{Ren, He, Girshick,\ \BBA\ Sun}{Ren
  et~al.}{2015}]{ren2015faster}
Ren, S., He, K., Girshick, R., \BBA\ Sun, J. \BBOP2015\BBCP.
\newblock \BBOQ Faster r-cnn: Towards real-time object detection with region
  proposal networks\BBCQ\
\newblock In {\Bem Advances in neural information processing systems}, \BPGS\
  91--99.

\bibitem[\protect\BCAY{Rohrbach, Rohrbach, Hu, Darrell,\ \BBA\
  Schiele}{Rohrbach et~al.}{2016}]{rohrbach2016grounding}
Rohrbach, A., Rohrbach, M., Hu, R., Darrell, T., \BBA\ Schiele, B.
  \BBOP2016\BBCP.
\newblock \BBOQ Grounding of textual phrases in images by reconstruction\BBCQ\
\newblock In {\Bem European Conference on Computer Vision}, \BPGS\ 817--834.
  Springer.

\bibitem[\protect\BCAY{Ryu, Matsubara,\ \BBA\ Inagaki}{Ryu
  et~al.}{2006}]{ryu-etal-2006-simultaneous}
Ryu, K., Matsubara, S., \BBA\ Inagaki, Y. \BBOP2006\BBCP.
\newblock \BBOQ Simultaneous {E}nglish-{J}apanese spoken language translation
  based on incremental dependency parsing and transfer\BBCQ\
\newblock In {\Bem Proceedings of the {COLING}/{ACL} 2006 Main Conference
  Poster Sessions}, \BPGS\ 683--690, Sydney, Australia. Association for
  Computational Linguistics.

\bibitem[\protect\BCAY{Satija\ \BBA\ Pineau}{Satija\ \BBA\
  Pineau}{2016}]{satija2016simultaneous}
Satija, H.\BBACOMMA\  \BBA\ Pineau, J. \BBOP2016\BBCP.
\newblock \BBOQ Simultaneous machine translation using deep reinforcement
  learning\BBCQ\
\newblock In {\Bem ICML 2016 Workshop on Abstraction in Reinforcement
  Learning}.

\bibitem[\protect\BCAY{Specia, Frank, Sima'an,\ \BBA\ Elliott}{Specia
  et~al.}{2016}]{specia-etal-2016-shared}
Specia, L., Frank, S., Sima'an, K., \BBA\ Elliott, D. \BBOP2016\BBCP.
\newblock \BBOQ A shared task on multimodal machine translation and
  crosslingual image description\BBCQ\
\newblock In {\Bem Proceedings of the First Conference on Machine Translation},
  \BPGS\ 543--553, Berlin, Germany. Association for Computational Linguistics.

\bibitem[\protect\BCAY{Specia, Wang, Jae~Lee, Ostapenko,\ \BBA\
  Madhyastha}{Specia et~al.}{2021}]{specia2020read-spot-translate}
Specia, L., Wang, J., Jae~Lee, S., Ostapenko, A., \BBA\ Madhyastha, P.
  \BBOP2021\BBCP.
\newblock \BBOQ Read, spot and translate\BBCQ\
\newblock {\Bem Machine Translation}, {\Bem 35\/}(1), 145--165.

\bibitem[\protect\BCAY{Sulubacak, Caglayan, Gr{\"o}nroos, Rouhe, Elliott,
  Specia,\ \BBA\ Tiedemann}{Sulubacak et~al.}{2020}]{sulubacak2020multimodal}
Sulubacak, U., Caglayan, O., Gr{\"o}nroos, S.-A., Rouhe, A., Elliott, D.,
  Specia, L., \BBA\ Tiedemann, J. \BBOP2020\BBCP.
\newblock \BBOQ Multimodal machine translation through visuals and speech\BBCQ\
\newblock {\Bem Machine Translation}, {\Bem 34\/}(2), 97--147.

\bibitem[\protect\BCAY{Sutskever, Vinyals,\ \BBA\ Le}{Sutskever
  et~al.}{2014}]{sutskever2014sequence}
Sutskever, I., Vinyals, O., \BBA\ Le, Q.~V. \BBOP2014\BBCP.
\newblock \BBOQ Sequence to sequence learning with neural networks\BBCQ\
\newblock In {\Bem Advances in neural information processing systems}, \BPGS\
  3104--3112.

\bibitem[\protect\BCAY{Szegedy, Vanhoucke, Ioffe, Shlens,\ \BBA\ Wojna}{Szegedy
  et~al.}{2016}]{szegedy2016rethinking}
Szegedy, C., Vanhoucke, V., Ioffe, S., Shlens, J., \BBA\ Wojna, Z.
  \BBOP2016\BBCP.
\newblock \BBOQ Rethinking the inception architecture for computer vision\BBCQ\
\newblock In {\Bem Proceedings of the IEEE conference on computer vision and
  pattern recognition}, \BPGS\ 2818--2826.

\bibitem[\protect\BCAY{Tan\ \BBA\ Bansal}{Tan\ \BBA\
  Bansal}{2019}]{tan-bansal-2019-lxmert}
Tan, H.\BBACOMMA\  \BBA\ Bansal, M. \BBOP2019\BBCP.
\newblock \BBOQ {LXMERT}: Learning cross-modality encoder representations from
  transformers\BBCQ\
\newblock In {\Bem Proceedings of the 2019 Conference on Empirical Methods in
  Natural Language Processing and the 9th International Joint Conference on
  Natural Language Processing (EMNLP-IJCNLP)}, \BPGS\ 5100--5111, Hong Kong,
  China. Association for Computational Linguistics.

\bibitem[\protect\BCAY{Vaswani, Shazeer, Parmar, Uszkoreit, Jones, Gomez,
  Kaiser,\ \BBA\ Polosukhin}{Vaswani et~al.}{2017}]{vaswani2017attention}
Vaswani, A., Shazeer, N., Parmar, N., Uszkoreit, J., Jones, L., Gomez, A.~N.,
  Kaiser, {\L}., \BBA\ Polosukhin, I. \BBOP2017\BBCP.
\newblock \BBOQ Attention is all you need\BBCQ\
\newblock In {\Bem Advances in neural information processing systems}, \BPGS\
  5998--6008.

\bibitem[\protect\BCAY{Wang\ \BBA\ Specia}{Wang\ \BBA\
  Specia}{2019}]{wang-specia-2019}
Wang, J.\BBACOMMA\  \BBA\ Specia, L. \BBOP2019\BBCP.
\newblock \BBOQ Phrase localization without paired training examples\BBCQ\
\newblock In {\Bem Proceedings of the IEEE/CVF Internaitonal Conference on
  Computer Vision (ICCV)}, Seoul, South Korea. {IEEE}.

\bibitem[\protect\BCAY{Wang, Li, Xiao, Zhu, Li, Wong,\ \BBA\ Chao}{Wang
  et~al.}{2019}]{wang-etal-2019-learning-deep}
Wang, Q., Li, B., Xiao, T., Zhu, J., Li, C., Wong, D.~F., \BBA\ Chao, L.~S.
  \BBOP2019\BBCP.
\newblock \BBOQ Learning deep transformer models for machine translation\BBCQ\
\newblock In {\Bem Proceedings of the 57th Annual Meeting of the Association
  for Computational Linguistics}, \BPGS\ 1810--1822, Florence, Italy.
  Association for Computational Linguistics.

\bibitem[\protect\BCAY{Williams}{Williams}{1992}]{williams1992simple}
Williams, R.~J. \BBOP1992\BBCP.
\newblock \BBOQ Simple statistical gradient-following algorithms for
  connectionist reinforcement learning\BBCQ\
\newblock {\Bem Machine learning}, {\Bem 8\/}(3-4), 229--256.

\bibitem[\protect\BCAY{Young, Lai, Hodosh,\ \BBA\ Hockenmaier}{Young
  et~al.}{2014}]{young-etal-2014-image}
Young, P., Lai, A., Hodosh, M., \BBA\ Hockenmaier, J. \BBOP2014\BBCP.
\newblock \BBOQ From image descriptions to visual denotations: New similarity
  metrics for semantic inference over event descriptions\BBCQ\
\newblock {\Bem Transactions of the Association for Computational Linguistics},
  {\Bem 2}, 67--78.

\bibitem[\protect\BCAY{Zheng, Zheng, Ma,\ \BBA\ Huang}{Zheng
  et~al.}{2019}]{zheng2019simpler}
Zheng, B., Zheng, R., Ma, M., \BBA\ Huang, L. \BBOP2019\BBCP.
\newblock \BBOQ Simpler and faster learning of adaptive policies for
  simultaneous translation\BBCQ\
\newblock In {\Bem Proceedings of the 2019 Conference on Empirical Methods in
  Natural Language Processing and the 9th International Joint Conference on
  Natural Language Processing (EMNLP-IJCNLP)}, \BPGS\ 1349--1354.

\bibitem[\protect\BCAY{Zhou, Cheng, Lee,\ \BBA\ Yu}{Zhou
  et~al.}{2018}]{zhou-etal-2018-visual}
Zhou, M., Cheng, R., Lee, Y.~J., \BBA\ Yu, Z. \BBOP2018\BBCP.
\newblock \BBOQ A visual attention grounding neural model for multimodal
  machine translation\BBCQ\
\newblock In {\Bem Proceedings of the 2018 Conference on Empirical Methods in
  Natural Language Processing}, \BPGS\ 3643--3653, Brussels, Belgium.
  Association for Computational Linguistics.

\end{thebibliography}
\bibliographystyle{theapa}

\end{document}